\documentclass[journal]{IEEEtran}

\usepackage{multirow}
\usepackage{xcolor}
\usepackage{graphicx,epstopdf}
\usepackage{amsmath,amsthm,amssymb,amsfonts}
\usepackage{tabularx,pdfcomment}
\usepackage{graphicx,color,multirow,amsmath}
\usepackage{cite}
\usepackage{makecell}

\usepackage{subfigure}
\usepackage{hyperref}
\usepackage{url}
\usepackage{cases}
\usepackage{bbm}
\usepackage{bm}
\usepackage{booktabs} 

\usepackage{algpseudocode}  
\usepackage{algorithmicx}  
\usepackage{algorithm}

\usepackage[square, comma, sort&compress, numbers]{natbib}

\theoremstyle{plain} 
\theoremstyle{plain} 
\theoremstyle{plain} 
\theoremstyle{plain} 
\theoremstyle{plain} 
\theoremstyle{plain}

\renewcommand\appendixname{Appendix}

\begin{document}
\title{Unsupervised Seismic Footprint Removal \textcolor{black}{With Physical Prior Augmented Deep Autoencoder}}

\author{Feng~Qian,~\IEEEmembership{Member,~IEEE,}
		 Yuehua~Yue,
        Yu~He,
        Hongtao~Yu, 
        Yingjie~Zhou,~\IEEEmembership{Member,~IEEE,}\\
        Jinliang~Tang,
        and~Guangmin~Hu,~\IEEEmembership{Member,~IEEE}
 
\thanks{Feng~Qian is with the School of Information and Communication Engineering and the Center for Information Geoscience, University of Electronic Science and Technology of China, Chengdu, 611731 China (e-mail: fengqian@uestc.edu.cn).}
\thanks{Yuehua~Yue, Yu~He, and Guangmin~Hu are all with the School of Resource and Environment and the Center for Information Geoscience, University of Electronic Science and Technology of China, Chengdu, 611731 China (e-mail:~3341452205@qq.com;~1141448613@qq.com;~hgm@uestc.edu.cn).}
\thanks{Yingjie~Zhou is with the College of Computer Science, Sichuan University (SCU), Chengdu 610065, China (e-mail: yjzhou@scu.edu.cn).}
\thanks{Jinliang~Tang is with the Sinopec Geophysical Research Institute (SGRI), Institute of Seismic Imaging Technology, Nanjing 211103, China (e-mail: tangjl.swty@sinopec.com).}
}

\markboth{SUBMITTED TO IEEE TRANSACTIONS ON GEOSCIENCE AND REMOTE SENSING}
{QIAN \MakeLowercase{\textit{et al.}}: Ground Truth-Free Footprint Suppression}

\maketitle
\begin{abstract}
Seismic acquisition footprints appear as stably faint and dim structures and emerge fully spatially coherent, causing inevitable damage to useful signals during the suppression process. Various footprint removal methods, including filtering and sparse representation (SR), have been reported to attain promising results for surmounting this challenge. However, these methods, e.g., SR, rely solely on the handcrafted image priors of useful signals, which is sometimes an unreasonable demand if complex geological structures are contained in the given seismic data. As an alternative, this article proposes a footprint removal network (dubbed FR-Net) for the unsupervised suppression of acquired footprints without any assumptions regarding valuable signals. The key to the FR-Net is to design a unidirectional total variation (UTV) model for footprint acquisition according to the intrinsically directional property of noise. By strongly regularizing a deep convolutional autoencoder (DCAE) using the UTV model, our FR-Net transforms the DCAE from an entirely data-driven model to a \textcolor{black}{prior-augmented} approach, inheriting the superiority of the DCAE and our footprint model. Subsequently, the complete separation of the footprint noise and useful signals is projected in an unsupervised manner, specifically by optimizing the FR-Net via the backpropagation (BP) algorithm. We provide qualitative and quantitative evaluations conducted on three synthetic and field datasets, demonstrating that our FR-Net surpasses the previous state-of-the-art (SOTA) methods.

\end{abstract}


\IEEEpeerreviewmaketitle

\section{Introduction}
\label{sec:Intro}
\IEEEPARstart{A}{cquisition} footprints are extensively perceived as anomalous and periodic amplitude stripes that are related to the acquisition geometry rather than geology, always appearing to cross the time slices or horizon slices of 3-D seismic data~\cite{chopra2000acquisition}. The leading causes of footprint noise are diverse and occur throughout the data acquisition and processing procedures, but two crucial aspects, including acquisition design and accentuation during processing, may play a role~\cite{hill1999acquisition}--\cite{li2008footprint}. The presence of footprint noise makes seismic images unclear and may cause confusion when identifying and analyzing the true subsurface response, which undoubtedly boosts uncertainty in seismic image interpretation~\cite{marfurt1998suppression},~\cite{cooper2010seismic}. As a result, footprint removal is a tremendously supreme yet inseparable step in seismic data processing~\cite{alali2018attribute},~\cite{sahai2006use} and is also the focus of this article.

Great efforts have been made by the geophysics community to boost the development of footprint noise removal approaches in recent decades ~\cite{chopra2000acquisition},~\cite{gulunay19943d}--\cite{gomez2020footprint}. Notably, a surplus of vital approaches has been advanced in the seismic literature, and they are roughly grouped into two categories: filtering~\cite{chopra2000acquisition},~\cite{gulunay19943d}--\cite{fehmers2003fast} and sparse representation (SR) methods~\cite{yu2017attenuation}--\cite{gomez2020footprint}. Filtering methods are the most extensively used techniques and can be divided into two classes: wavenumber/frequency-domain and time-domain filtering. The first class, initially used by Gulunay~\cite{gulunay19943d}, handles footprints as spatially periodic noise, resulting in suppression attempts in the wavenumber space; examples include classic wavenumber filtering~\cite{gulunay19943d},~\cite{soubaras2002attenuation}, adaptive wavenumber filtering~\cite{falconer2008attribute}, spectral notch
filtering~\cite{gulunay1999acquisition}, two-pass frequency-wavenumber filtering~\cite{chopra2000acquisition}, and wavenumber notch filtering~\cite{drummond2000adapting}. The second class is primarily appropriate for random noise abatement but can also be applied to footprint removal. Typical methods include singular value decomposition (SVD)-based~\cite{al2005acquisition}--\cite{chen2016simultaneous} and structure-oriented filtering~\cite{fehmers2003fast}. However, in case studies, conflict arises from the fact that filtering technology attempts to achieve significant footprint removal while ensuring that the true amplitude is not influenced.

Compared to the wavenumber/frequency-domain approaches, SR has been suggested to achieve more compact representations of both clear seismic images and footprints, which allows them to be better separated in the sparsifying domain~\cite{yu2017attenuation}--\cite{gomez2020footprint}. Depending on how correctly the utilized dictionaries are formed, SR methods can be divided into analytical sparsifying transforms and data-driven sparse coding (SC) approaches. In the analytical sparsifying transform case, it is assumed that noise suppression is based on the calculation of sparse coefficients for footprints and clear images with different distribution laws in a given analytical transform basis; representative techniques include the wavelet transform~\cite{alali2018attribute}, the 3-D complex wavelet transform~\cite{yu2017attenuation}, the stationary wavelet transform (SWT)~\cite{chen2012adaptive},~\cite{cvetkovic20072d}, basis pursuit~\cite{zhang2009footprint}, and the curvelet transform~\cite{gorszczyk2015enhancing}. Concretely, based on the analysis of the energy distribution laws of coefficients, the complete separation of footprints and clear seismic images is achieved by performing threshold or filtering operations on the sparse coefficients. Entirely unlike analytical sparsifying transforms, considering that the atoms of the learned dictionary are found to represent clear images and footprint patterns~\cite{chen2021statistics},~\cite{liu2021dictionary}, SC-based denoising methods naturally deal with these learned atoms and ignore the role of sparse coefficients. In particular, SC methods consider a footprint suppression problem as a binary pattern classification~\cite{chen2021statistics}--\cite{gao2016local} or filtering task~\cite{gomez2020footprint} for all learned atoms, which enables the separation of footprints and clear images at the atom level. However, although the existing methods have achieved great progress in footprint removal, their performance relies on handcrafted priors and may be degraded if the input seismic image has unexpected properties beyond the model assumptions.

Inspired by the overwhelming success of deep learning (DL) as a data processing technique, DL-based methods have emerged for seismic image denoising in recent years~\cite{yu2019deep}--\cite{zhao2018low}. In contrast with conventional noise cancellation methods, a DL-based denoising method permanently seeks to intuitively learn an underlying mapping between a noisy seismic image and the corresponding clean image without any handcrafted assumptions~\cite{yu2019deep}--\cite{qian2021dtae}. In this case, DL-based denoising methods tend to concentrate on random noise and aim to minimize the mean squared error (MSE) induced over the training set using various DL models. Typical DL networks include convolutional neural networks (CNNs)~\cite{yu2019deep}--\cite{yang2021deep}, 3-D denoising CNNs (3D-DnCNNs)~\cite{liu2018random},~\cite{liu2019poststack}, deep convolutional autoencoders (DCAEs)~\cite{zhang2019unsupervised},~\cite{saad2020deep}, 3-D DCAEs~\cite{yang2021unsupervised},~\cite{saad2021fully}, and tensor DL (TDL) methods~\cite{qian2022ground},~\cite{qian2021dtae}. Other areas of open research include utilizing DL for the suppression of erratic noise~\cite{qian2022unsupervised}, coupled noise~\cite{zhao2020distributed}--\cite{zhao2022coupled}, swell noise~\cite{zhao2019swell},~\cite{farmani2020application}, ground-roll noise~\cite{li2018deep}--\cite{yuan2020ground}, and desert noise~\cite{ma2019deep},~\cite{zhao2018low}. However, no DL methods are currently deployed in the geophysics community to assist with suppressing acquired footprints, since the collection of footprints and their clean counterparts is a nontrivial task.


To ameliorate this issue, \textcolor{black}{we propose a footprint removal network (FR-Net) that is directly driven by footprint's physical prior and related model, thus allows a more complete unsupervised decoupling of footprints and reflected signals  compared to prevailing removal methods. The superior performance of our suggested FR-Net can be attributed to the following characteristics.} 1) \textit{\textcolor{black}{Physical prior augmented DL}}: Pure DL methods do not work for footprint noise; the well-equipped MSE loss pertains to the Gaussian errors incurred during error propagation, causing it to fail to remove footprints. Based on \textcolor{black}{the footprint's physical nature and its modeling}, our FR-Net transforms DL from fully data-driven to \textcolor{black}{physical prior augmented}, enabling it to inherit the superiority of both DL and our footprint model and thus to achieve outstanding performance. 2) \textit{Explainable}: Efforts are made in this article to produce explainable deep neural network (xDNN) or explainable machine learning (XML) techniques, with the goal of providing a stronger descriptive approach for the FR-Net as well as additional prior information for users, hence enhancing the insights obtained regarding footprint noise. 3) \textit{Ground truth-free}: A somewhat surprising feature of our FR-Net model is that, despite the use of a DL model, the FR-Net can effectively remove footprint noise in an unsupervised manner, which does not require the acquisition of footprints and their clean counterparts.

\textcolor{black}{Relying on the advantages listed above, it is clarified that the FR-Net is a promising method for conducting footprint removal on noisy 3-D seismic data, which is the primary purpose of this article. In contrast with current footprint suppression methods, three substantial evolutions make our FR-Net model technically feasible, and they are summarized below as our main contributions.}

\begin{itemize}
\item [1)] \textit{Footprint modeling:} The key to the FR-Net is to design a footprint separation model (i.e., a unidirectional total variation (UTV) model) according to the intrinsically directional property of footprint noise (x-line axis or y-line axis). From the above literature survey, it can be seen that our footprint model is quite distinct from all current removal methods.

\item [2)] \textit{\textcolor{black}{Physical prior augmented} DL:} Inspired by~\cite{zhang2019unsupervised},~\cite{saad2020deep},~\cite{qian2022unsupervised}, we seek to develop a DL model that can automatically describe the ideal image behavior hidden in contaminated seismic data. Then, strongly regularizing the DL approach using the UTV model produces an ultimate FR-Net for \textcolor{black}{thoroughly decoupling footprints and reflected signals.} To the best of our knowledge, this article is the first work that uses DL for unsupervised footprint removal.

\item [3)] \textit{Solution:} The complete separation of footprint noise and useful signals is achieved when the FR-Net can be trained by the derived backpropagation (BP) algorithm, which is based on the classic BP algorithm. Additionally, this BP deduction process may supply insight into the \textcolor{black}{physical prior augmented} nature of the derivative calculations during the training of the FR-Net. Nevertheless, in the specific implementation, the derivatives of this new cost function are processed automatically by TensorFlow~\cite{abadi2016tensorflow}.

\item [4)] \textit{Geophysics community:} \textcolor{black}{As footprints are typical coherent noise, the present \textcolor{black}{physical prior augmented} idea is unprecedented with respect to the aforementioned seismic data denoising process, particularly coherent noise suppression. It will undoubtedly pertain to our future research direction and dramatically enrich the theory of seismic noise abatement itself.}

\end{itemize}
Extensive experimental results obtained on three public datasets demonstrate the supremacy and efficacy of the proposed FR-Net in comparison with existing state-of-the-art (SOTA) methods.

The rest of this paper is organized as follows. Section \ref{sec:pre} gives some background on the DCAE and total variation (TV) model used throughout this article. The problem formulation and modeling-related issues are discussed in Section \ref{sec:form}. Section \ref{Sec:method} completely describes the proposed work. Section \ref{Sec:examp} present the experimental results and an evaluation to prove the efficacy of the proposed FR-Net. Finally, Section \ref{Sec:conclusion} concludes our work.

\begin{figure*}[t!]

	\vspace{-0.4cm}
	\centering
	\includegraphics[width=\textwidth]{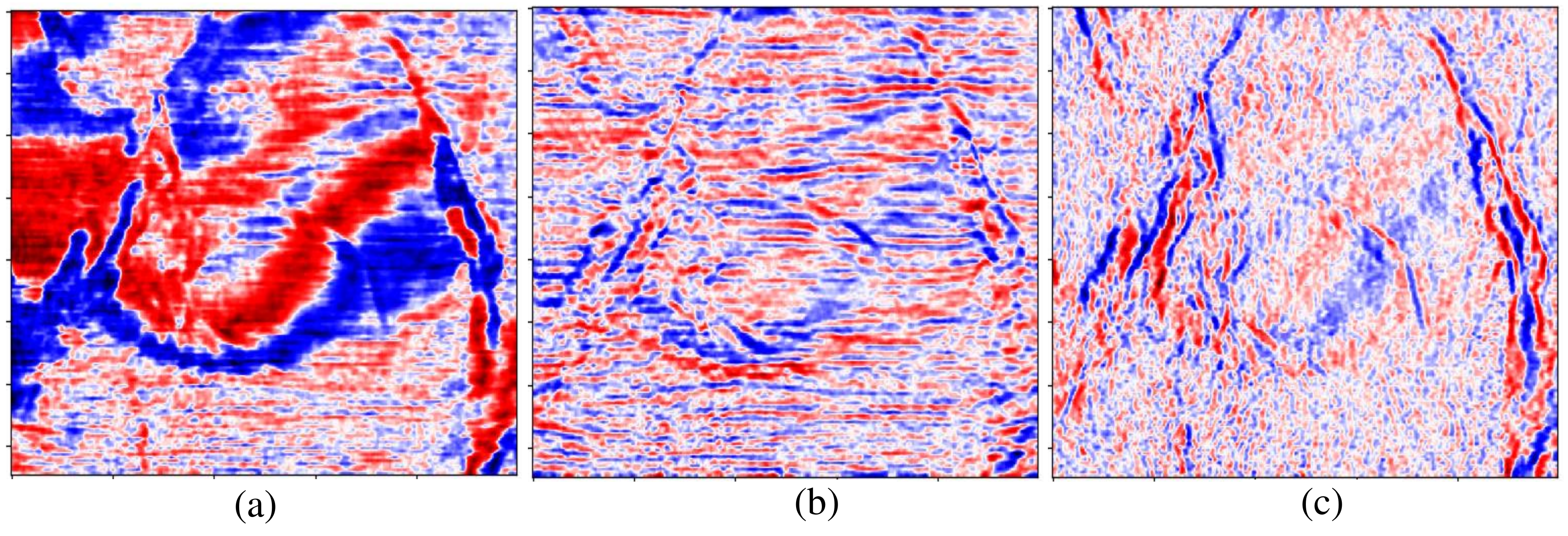}
	\caption{(a) 
Seismic image with footprint noise. 
(b) Horizontal derivative. (c) Vertical derivative.}
\label{Fig:motiv}
\end{figure*}

\section{Background}
\label{sec:pre}
For the convenience of the reader's understanding of this work, this part briefly reviews two essential topics to provide the necessary preliminaries for presenting the proposed FR-Net. Specifically, some rudimentary knowledge concerning DCAEs is revisited in Section~\ref{Sec:ae}. Then, TV is succinctly described in Section~\ref{Sec:TV}.

\subsection{DCAE}
\label{Sec:ae}
A DCAE is an end-to-end model that consists of two components: an encoder with multiple layers that first encodes an input seismic image to lower-dimensional latent features and a decoder that decodes the latent representation back into a seismic image~\cite{hinton1993autoencoders}. Suppose that the input $\{\mathbf{X}_{n}\}_{n=1}^N$ is known; the hidden representations for each layer can be obtained by the following:

\begin{equation}\textcolor{black}{\begin{array}{l}
\mathbf{X}_{n,\ell} = \mathbf{h}\left(\mathbf{X}_{n,\ell-1}\right)=\sigma\left(\mathbf{W}_{\ell-1} \mathbf{X}_{n,\ell-1}+\mathbf{b}_{\ell}\right) \\
{\mathbf{g}}\left(\mathbf{X}_{n,\ell-1}\right)=\sigma\left(\mathbf{W}_{\ell-1}^{\top} \mathbf{h}\left(\mathbf{X}_{n,\ell-1}\right)+\mathbf{b}_{\ell-1}\right),\quad \ell\in[K],
\end{array}}
\label{equ:ae2}
\end{equation}
where $(\cdot)^{\top}$ denotes the transpose of a given matrix and $N$ signifies the number of input sample couples. Since the decoder is designed as the reverse of the encoder, one can feed $\mathbf{X}_{n,\ell-1}$ into the decoder and compare the output $\mathbf{X}_{n,\ell}$ of the latter with the original images by minimizing the subsequent reconstruction error cost.



\begin{equation}\textcolor{black}{\mathcal{L}_{\mathrm{MSE}}(\mathbf{Y}_{n},\mathbf{X}_{n,\ell-1};\theta)=\arg \min _{\theta} \frac{1}{N}\sum_{n=1}^{N} \frac{1}{2}\left\|\mathbf{Y}_{n}-\mathbf{X}_{n,\ell}\right\|_{\mathrm{F}}^{2},}
\label{equ:MSEobj}
\end{equation}
\textcolor{black}{where $\theta = \{\mathbf{W}_{\ell},\mathbf{b}_{\ell}\}_{\ell=1}^{K}$ denotes the parameters \textcolor{black}{of the AE model.}}

\subsection{TV}
\label{Sec:TV}
The TV model was initially suggested by Rudin et al.~\cite{rudin1992nonlinear} and was formed to tackle grayscale image denoising issues due to its ability to retain edge information and efficiently facilitate piecewise smoothness. Later, the extrusion of the image denoising technique by the TV model became increasingly critical in view of its superior performance, making it a natural choice for extending seismic image noise lessening.

Formally, for a matrix $\mathbf{X} \in \mathbb{R}^{n_1 \times n_2}$, its derivative matrices $\nabla_{x} \mathbf{X} \in \mathbb{R}^{(n_1-1) \times n_2}$ and $\nabla_{y} \mathbf{X} \in \mathbb{R}^{n_1 \times(n_2-1)}$ are defined as

\begin{equation}
\left\{\begin{aligned}
\nabla_{x} \mathbf{X}(i, j) &=\mathbf{X}(i+1, j)-\mathbf{X}(i, j) \\
\nabla_{y} \mathbf{X}(i, j) &=\mathbf{X}(i, j+1)-\mathbf{X}(i, j), 
\end{aligned}\right.
\end{equation}
where $\mathbf{X}(i, j)$ denotes the $(i, j)$th element of $\mathbf{X}$, and $\nabla_{x}$ and $\nabla_{y}$ denote the first-order vertical and horizontal derivative operators, respectively. The TV of $\mathbf{X}$ is written as
\begin{equation}
\|\mathbf{X}\|_{\mathrm{TV}}=\left\|\nabla_{x} \mathbf{X}\right\|_{{1}}+\left\|\nabla_{y} \mathbf{X}\right\|_{{1}}.
\label{}
\end{equation}

After determining the TV norm of image denoising, we can treat it as a prior and incorporate it to constrain the maximum a posteriori probability (MAP) estimation, which is defined as follows:

\begin{equation}
\hat{\mathbf{X}}=\underset{\mathbf{X} \in \mathbb{R}^{m \times n}}{\arg \min }~\|\mathbf{Y}-\mathbf{X}\|_{F}^{2}+\tau\|\mathbf{X}\|_{\mathrm{TV}},
\end{equation}
where $\tau$ denotes the noise level of the Gaussian random noise.
\begin{figure*}[t!]

	\vspace{-0.4cm}
	\centering
	\includegraphics[width=\textwidth]{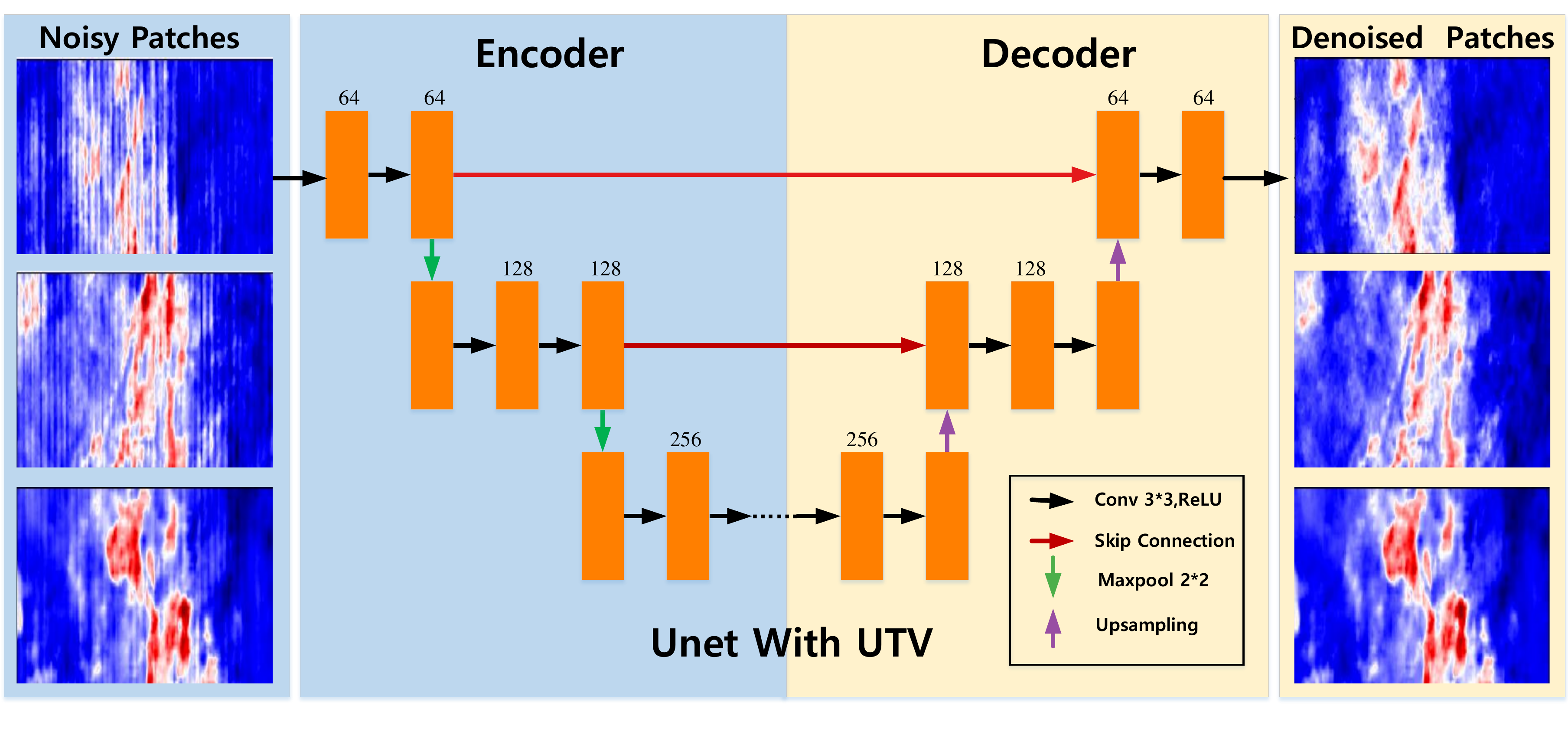}
	\caption{Schematic of our FR-Net model concept.}
\label{Fig:Schem} 
\end{figure*}

\section{Problem Formulation and Modeling}
\label{sec:form}
This section proposes a \textcolor{black}{physical prior augmented} DL approach for footprint removal. To elaborate, we begin by stating and formulating the footprint removal problem in Section~\ref{subsec:prob}. Then, the key to such a \textcolor{black}{physical prior augmented} approach is establishing a UTV model for footprints in Section~\ref{subsec:mod}. Finally, strongly regularizing the DL approach using the UTV model produces an ultimate FR-Net for thoroughly removing footprints in Section~\ref{subsec:seismod}.

\subsection{Problem Formulation}
\label{subsec:prob}
{As noted in~\cite{chopra2000acquisition}, acquired footprints refer to linear spatial grid patterns that are evident on 3-D seismic time or horizon slices. Therefore, as depicted in Fig.~\ref{Fig:motiv}(a), the noisy data $\mathbf{Y}$ are treated as a mixture of the ground-truth image $\mathbf{X}$, footprint noise $\mathbf{F}$, and random noise $\mathbf{N}$. Furthermore,
given that acquired footprint noise and random noise are additive~\cite{gomez2020footprint}, the deterioration process caused by footprints and random noise can be stated as follows.

\begin{equation}
\mathbf{Y}=\mathbf{X}+\mathbf{F}+\mathbf{N}.
\label{equ:sys}
\end{equation}
Formally, our objective is to infer the latent clear seismic image $\mathbf{X}$ from the noisy image $\mathbf{Y}$ by minimizing the following equation.

\begin{equation}
\min _{\mathbf{X}, \mathbf{F}}~\frac{1}{2}\|\mathbf{X}+\mathbf{F}-\mathbf{Y}\|_{F}^{2}.
\end{equation}
Mathematically, since the task is a typical ill-posed inverse problem, the key is to design appropriate regularization terms to simultaneously remove both footprints $\mathbf{F}$ and random noise $\mathbf{N}$ as follows:

\begin{equation}
\min _{\mathbf{X}, \mathbf{F}}~\frac{1}{2}\|\mathbf{X}+\mathbf{F}-\mathbf{Y}\|_{F}^{2}+\tau P(\mathbf{X})+\lambda P(\mathbf{F}),
\label{equ:inv}
\end{equation}
where $P(\mathbf{X})$ and $P(\mathbf{F})$ denote the priors concerning the clear seismic image and footprint components, respectively, and both $\tau$ and $\lambda$ denote regularization parameters. To date, the two most typically used techniques, explicit filtering~\cite{chopra2000acquisition},~\cite{gulunay19943d}--\cite{fehmers2003fast} and SR~\cite{yu2017attenuation}--\cite{gomez2020footprint}, have been advanced for footprint suppression. From these methods, it can be seen that the key to footprint noise reduction is to build an appropriate model for clean images and footprints, which significantly facilitates the separation of the two
components.

\subsection{Footprint Modeling}
\label{subsec:mod}
The most important aspect of footprint removal is the modeling of footprint noise, and the main existing methods treat footprints as spatially periodic noise that is sparse in the wavenumber or time-frequency space~\cite{gulunay19943d}--\cite{drummond2000adapting}. \textcolor{black}{Similar to the existing methods, we blatantly take advantage of another property: the spatial pattern of a footprint is regular linear stripes or grid patterns in terms of time slices~\cite{gulunay19943d}--\cite{drummond2000adapting}.} Making use of this linear feature, a TV technique is naturally chosen for the footprint removal task because of its edge preservation property or the spatial local smoothness~\cite{rudin1992nonlinear}. More importantly, TV can easily be imposed on the loss of the DL model in the form of a regular term~\cite{qian2022unsupervised},~\cite{guo2019toward}.

However, as depicted in Fig.~\ref{Fig:motiv}(a), it is highly visible that a footprint has a clear directional signature. For this, we argue that classic TV is inappropriate for footprint removal due to its noncompliance with the directional characteristics of footprints. As an illustration, Figs.~\ref{Fig:motiv}(b) and (c) depict the horizontal and vertical unidirectional gradients of the degraded seismic image, respectively. In Figs.~\ref{Fig:motiv}(b) and (c), the footprints have an enormous impact on the vertical gradient across the footprints, but they have no effect whatsoever on the gradient that is detected along the footprints. This fact compels us to limit the gradient between footprints while retaining the gradient along footprints. To accomplish this, one can utilize the UTV model initially developed by Bouali and Ladjal~\cite{bouali2011toward} that exploits the directional characteristics of footprints. The formalization of the UTV model is expressed as

\begin{figure*}[t!]

	\vspace{-0.4cm}
	\centering
	\includegraphics[width=\textwidth]{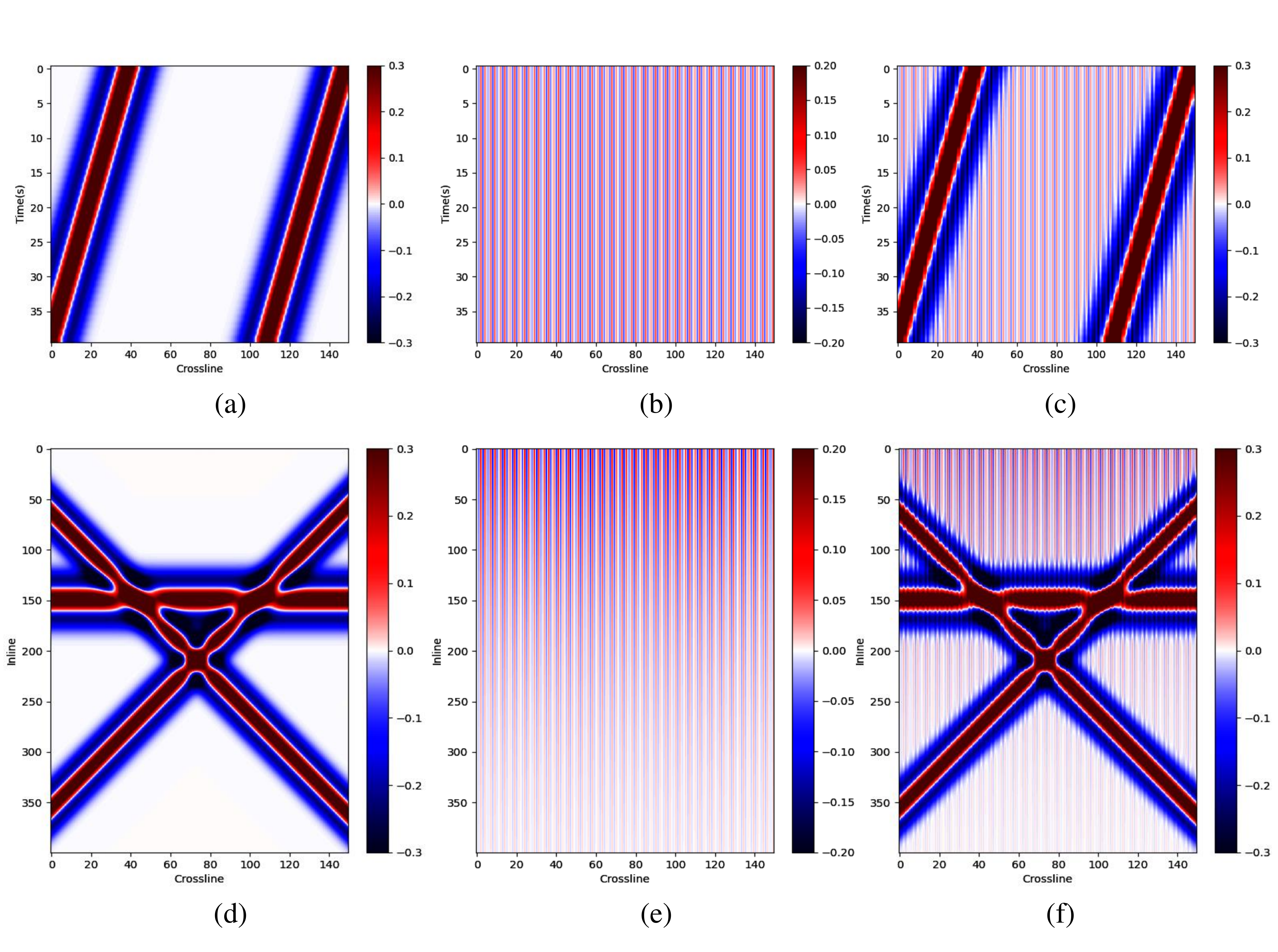}
	\caption{Synthetic example. (a and d) Clean seismic image. (b and e) Footprint. (c and f) Noisy seismic image.}
\label{Fig:syn1}
\end{figure*}

\begin{equation}
\mathcal{L}_{\mathrm{UTV}}(\mathbf{X})=\lambda_{1}\left\|\nabla_{x} \mathbf{X}\right\|_{1}+\lambda_{2}\left\|\nabla_{y}(\mathbf{Y}-\mathbf{X})\right\|_{1},
\label{equ:UTVobj}
\end{equation}
where $\lambda_{1}$ and $\lambda_{2}$ denote the tradeoff parameters. $\nabla_{x}$ and $\nabla_{y}$ represent the linear first-order difference operator in the x-line direction and y-line direction, respectively. \textcolor{black}{In this UTV model, the first term penalizes the $\ell_1$-norm of the gradient along the perpendicular direction of the footprints to remove footprints noise, while the second term enforces the $\ell_1$ norm constraints on the difference between the gradient along the direction of the footprints to reveal or separate underlying footprints.} Substituting equation (\ref{equ:UTVobj}) into (\ref{equ:inv}), one can obtain a robust stripe removal method by minimizing an objective function that includes unidirectional variations as follows:

\begin{equation}
\min _{\mathcal{X}}~\frac{1}{2}\|\mathbf{X}-\mathbf{Y}\|_{F}^{2}+\tau P(\mathbf{X})+\lambda_{1}\left\|\nabla_{x} \mathbf{X}\right\|_{1}+\lambda_{2}\left\|\nabla_{y}(\mathbf{Y}-\mathbf{X})\right\|_{1}.
\label{equ:mod1}
\end{equation}

To solve equation (\ref{equ:mod1}) with respect to $\mathbf{X}$, it is necessary to construct a model for $\mathbf{X}$ that imposes additional constraints on the highly ill-posed system equation.

\subsection{\textcolor{black}{Physical Prior Augmented} DL Modeling}
\label{subsec:seismod}
Based on the literature survey in Section~\ref{sec:Intro}, it is found that the current research on modeling $\mathbf{X}$ relies mainly on handcrafted priors. However, these prior assumptions do not always hold true for $\mathbf{X}$, and it is necessary to use DL to perform modeling in a data-driven way. Typically, a clear seismic image $\widehat{\mathbf{X}}$ can be modeled as a nonlinear autoregression in this situation:
\begin{equation}
\widehat{\mathbf{X}}=\boldsymbol{h}_{\Theta}(\mathbf{Y}).
\label{equ:dl}
\end{equation}
By conducting unsupervised training for a DNN on $\mathbf{X}$, one may approximate the denoising model $\boldsymbol{h}$ (parameterized by $\Theta$). Using this relation in equation (\ref{equ:dl}), we may rewrite equation (\ref{equ:MSEobj}) as:
\begin{equation}
\min _{\Theta}~\frac{1}{2}\|\boldsymbol{h}_{\Theta}(\mathbf{Y})-\mathbf{Y}\|^{2}.
\label{equ:dlmod}
\end{equation}
Afterward, by substituting equation (\ref{equ:dlmod}) into (\ref{equ:mod1}), one can achieve a final cost function for the incorrect footprint removal result, incorporating both the physical model of $\mathbf{F}$ and the data-driven model of $\mathbf{X}$, as follows.

\begin{equation}
\begin{aligned}
\min _{\Theta}~&\frac{1}{2}\left\|\boldsymbol{h}_{\Theta}(\mathbf{Y})-\mathbf{Y}\right\|_{2}^{2}+\lambda_{1}\left\|\nabla_{y} \boldsymbol{h}_{\Theta}(\mathbf{Y})\right\|_{1}\\
&+\lambda_{2}\left\|\nabla_{x}\left(\boldsymbol{h}_{\Theta}(\mathbf{Y})-\mathbf{Y}\right)\right\|_{1}.
\end{aligned}
\label{equ:tdlobj}
\end{equation}

From equation (\ref{equ:tdlobj}), it is surprising to find that the optimal parameters are changed from $\mathbf{X}$ to $\Theta$ to successfully derive a new DL model. An additional challenge involves accounting for the exclusive DL network structure that should be associated with the use of unsupervised footprint removal. }

\begin{figure*}[t!]

	\vspace{-0.4cm}
	\centering
	\includegraphics[width=\textwidth]{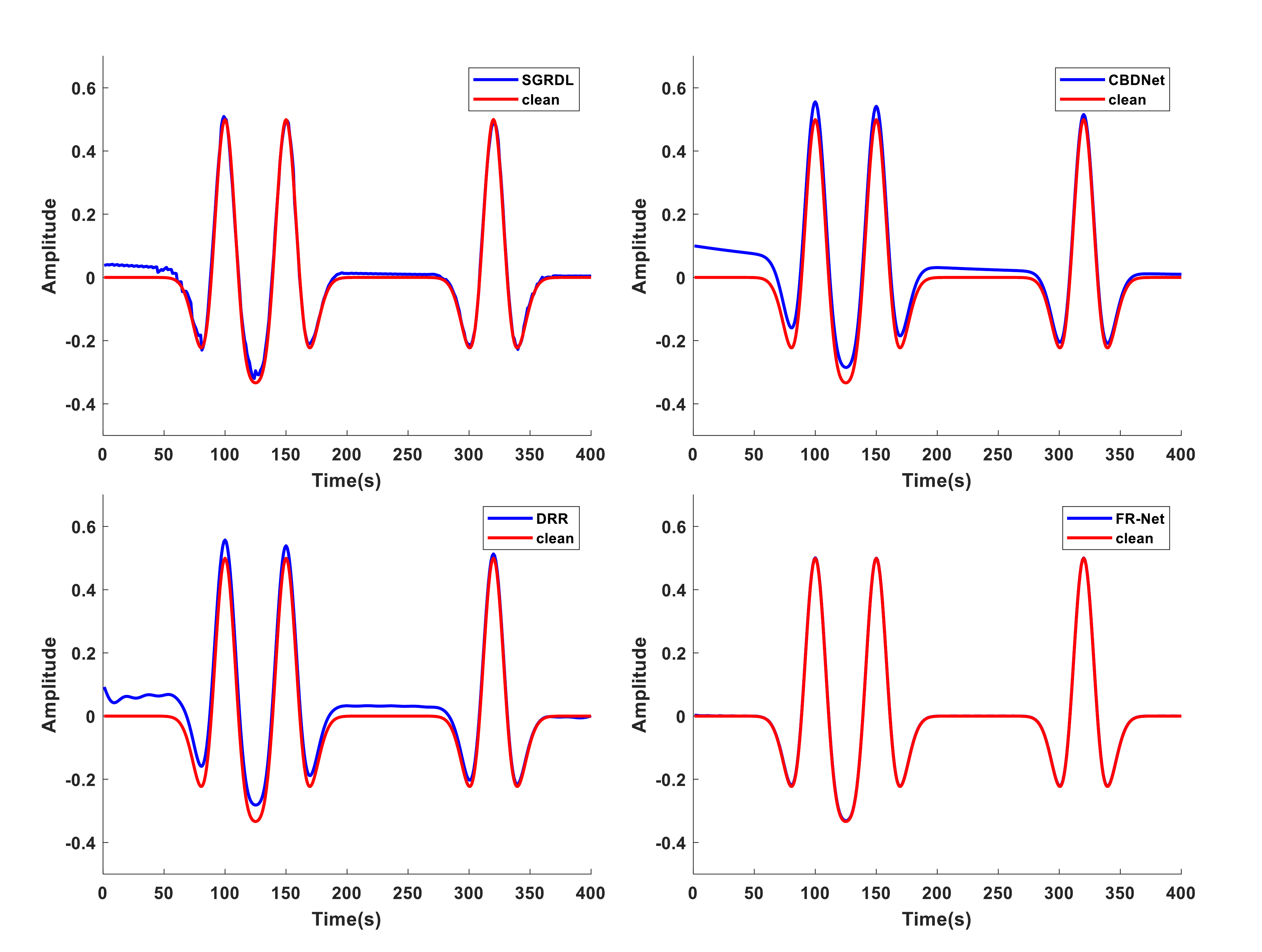}
	\caption{Results obtained on the synthetic dataset in the single-trace comparison.
}
\label{Fig:trace} 
\end{figure*}

\section{Methodology}
\label{Sec:method}
This section is separated into three distinct parts. Section~\ref{equ:utv} gives the proposed FR-Net model for footprint removal. Section~\ref{subsec:optim} provides the corresponding BP algorithm for determining the new model parameters. Section~\ref{sec:implement} presents the implementation details for the FR-Net.

\subsection{UTV Regularization for DCAEs}
\label{equ:utv}
As proven by ~\cite{saad2020deep},~\cite{yang2021unsupervised}, a DCAE can successfully denoise seismic images in an unsupervised manner, and it outperforms other model-based learning methods. As in our previous works~\cite{qian2022ground},~\cite{qian2022unsupervised}, one can use a DCAE to model clear image patches $\widehat{\mathbf{X}}$ and train a DL network to restore a clean image from a corrupted seismic image in a convenient form.

\begin{equation}
\widehat{\mathbf{X}}=\boldsymbol{h}_{\Theta}(\mathbf{Y})=\boldsymbol{D}_{\Theta}\left(\boldsymbol{E}_{\Theta}(\mathbf{Y})\right),
\label{equ:tdcae}
\end{equation}
where $\boldsymbol{D}_{\Theta}(\cdot)$ represents the encoder with $\Theta$. According to equation (\ref{equ:ae2}), $\boldsymbol{D}_{\Theta}(\cdot)$ is defined in detail as follows.

\begin{equation}
\mathbf{X}_{n, \ell}=\boldsymbol{E}_{\Theta}\left(\mathbf{X}_{n, \ell-1}\right)=\sigma\left(\mathbf{W}_{\ell-1} \mathbf{X}_{n, \ell-1}+\overrightarrow{\mathbf{B}}_{\ell}\right), \quad \ell \in[K].
\end{equation}

In contrast, $\boldsymbol{E}_{\Theta}(\cdot)$ represents the decoder that upsamples and maps the low-resolution code to a reconstruction of the original input, which is defined as

\begin{equation}
\boldsymbol{D}_{\Theta}\left(\mathbf{X}_{n, \ell-1}\right)=\sigma\left(\mathbf{W}_{\ell-1}^{\top}  \mathbf{h}\left(\mathbf{X}_{n, \ell-1}\right)+\overrightarrow{\mathbf{B}}_{\ell-1}\right), \quad \ell \in[K].
\end{equation}
Afterward, substituting equation (\ref{equ:tdcae}) into (\ref{equ:dlmod}), we obtain

\begin{equation}
\begin{aligned}
\mathcal{L}=\min _{\Theta}~&\frac{1}{2}\left\|\boldsymbol{D}_{\Theta}\left(\boldsymbol{E}_{\Theta}(\mathbf{Y})\right)-\mathcal{Y}\right\|_{2}^{2}+\lambda_{1}\left\|\nabla_{y} \boldsymbol{h}_{\Theta}(\mathbf{Y})\right\|_{1}\\
&+\lambda_{2}\left\|\nabla_{x}\left(\boldsymbol{h}_{\Theta}(\mathbf{Y})-\mathbf{Y}\right)\right\|_{1}.
\end{aligned}
\label{equ:FRobj}
\end{equation}
The footprint removal problem can be modeled with a UTV-regularized AE and solved by the BP algorithm, as in any conventional AE. However, for the $\mathcal{L}_{\mathrm{UTV}}$ regularity problem, it is more appropriate to solve the problem with this approach than the normal $\mathcal{L}_{\mathrm{MSE}}$, as their cost functions are quite diverse. Thus, it is essential to deduce a dedicated BP method to resolve the parameters $\Theta$ of equation (\ref{equ:FRobj}) by minimizing the new cost function (\ref{equ:FRobj}); its detailed derivation is described in Section~\ref{subsec:optim}.

\begin{figure*}[t!]

	\vspace{-0.4cm}
	\centering
	\includegraphics[width=\textwidth]{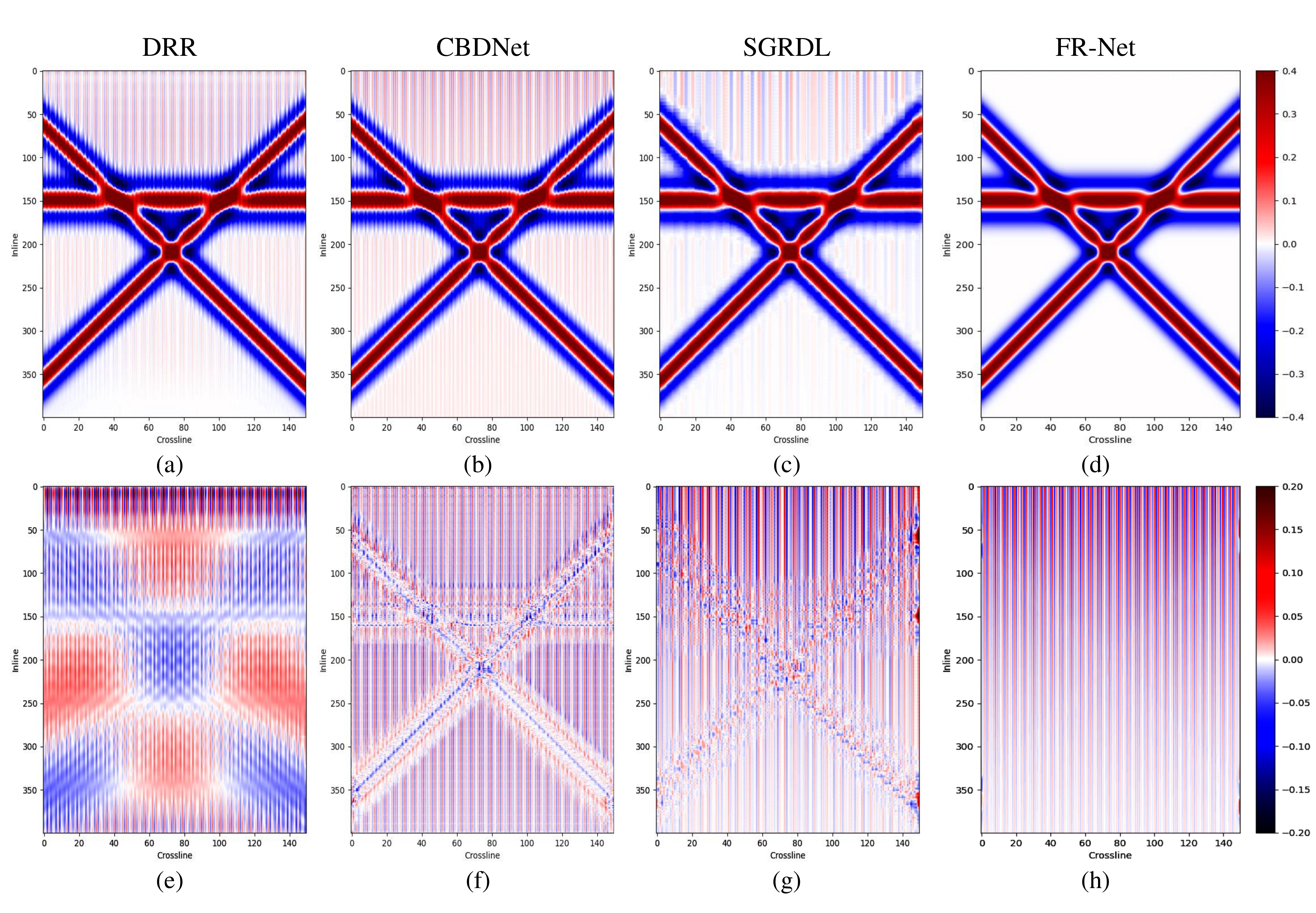}
	\caption{Results obtained on the synthetic dataset in the x-line section comparison. Footprint removal results are obtained using (a) the DRR method, (b) the CBDNet method, (c) the SGRDL method, and (d) the proposed FR-Net method. The corresponding residuals are obtained using
(e) the DRR method, (f) the CBDNet method, (g) the SGRDL method, and (h) the proposed FR-Net method.}
\label{Fig:syn2} 
\end{figure*}

\begin{figure*}[t!]

	\vspace{-0.4cm}
	\centering
	\includegraphics[width=\textwidth]{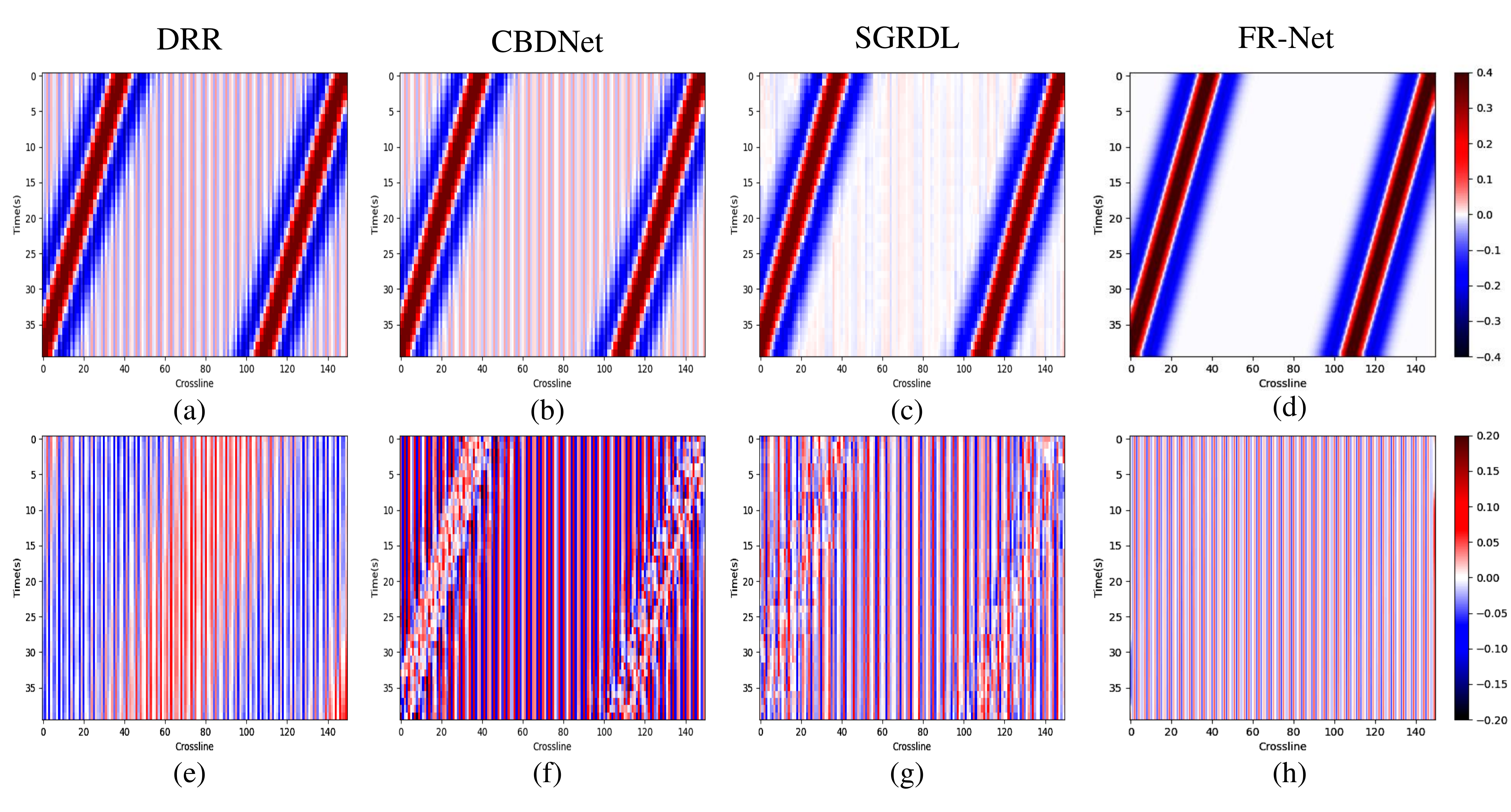}
	\caption{Results obtained on the synthetic dataset in the constant time slice comparison. Footprint removal results are obtained using (a) the DRR method, (b) the CBDNet method, (c) the SGRDL method, and (d) the proposed FR-Net method. The corresponding residuals are obtained using
(e) the DRR method, (f) the CBDNet method, (g) the SGRDL method, and (h) the proposed FR-Net method.}
\label{Fig:Syn3} 
\end{figure*}

\subsection{Optimization of the FR-Net}
\label{subsec:optim}

To explore the essential role of UTV regularization in footprint removal, we provide a straightforward derivation of the BP algorithm for the FR-Net in this section. Concretely, as an iterative optimization approach, gradient descent (GD) is used to gradually adjust the model parameters $\Theta$ by minimizing $\mathcal{L}\left(\mathbf{Y}_{n},\Theta\right)$ in (\ref{equ:FRobj}).
To follow the derivation, one can obtain a clearer idea about how GD modifies $\mathbf{W}_{\ell-1}$ and $\mathbf{B}_{\ell}$:

\begin{equation}
\begin{array}{r}
\mathbf{W}_{\ell-1} \leftarrow \mathbf{W}_{\ell-1}-\alpha \frac{\partial \mathcal{L}(\mathbf{W}, \mathbf{B})}{\partial \mathbf{W}_{\ell-1}}, 
\end{array}
\label{equ:wgd}
\end{equation}
\begin{equation}
\begin{array}{r}
\mathbf{B}_{\ell} \leftarrow \mathbf{B}_{\ell}-\alpha \frac{\partial \mathcal{L}(\mathbf{W}, \mathbf{B})}{\partial \mathbf{B}_{\ell}},
\end{array}
\end{equation}
where $\alpha$ denotes the learning rate, which is a hyperparameter that controls the step size per iteration. Typical GD algorithms proceed with alternate updates of $\mathbf{W}_{\ell}$ and $\mathbf{B}_{\ell}$. In particular, given the current estimate of $\mathbf{B}_{\ell}$, $\mathbf{W}_{\ell}$ can be readily updated by computing $\partial \mathcal{L} / \partial \mathbf{W}_{\ell-1}$ in equation (\ref{equ:wgd}), and $\partial \mathcal{L} / \partial \mathbf{W}_{\ell-1}$ can be performed with standard multiplicative rules as follows.

\begin{equation}
\label{equ:LW}
\delta \mathbf{W}_{\ell-1}=\frac{\partial \mathcal{L}}{\partial \mathbf{W}_{\ell-1}}=\frac{\partial \mathcal{L}_{\mathrm{MSE}}}{\partial \widehat{\mathbf{X}}_{n, \ell}} \frac{\partial \widehat{\mathbf{X}}_{n, \ell}}{\partial \mathbf{W}_{\ell-1}}+\frac{\partial \mathcal{L}_{\mathrm{UTV}}}{\partial \widehat{\mathbf{X}}_{n, \ell}} \frac{\partial \widehat{\mathbf{X}}_{n, \ell}}{\partial \mathbf{W}_{\ell-1}}.
\end{equation}

First, $\partial \mathcal{L}_{\mathrm{Welsch}} / \partial \widehat{\mathbf{X}}_{n, \ell}$ is found by taking the derivative of (\ref{equ:MSEobj}), which is:

\begin{equation}
\frac{\partial \mathcal{L}_{\mathrm{MSE}}}{\partial \widehat{\mathbf{X}}_{n, \ell}}=\mathbf{Z}.
\end{equation}
To simplify the presentation, we introduce the following definition: $\mathbf{Z}=\mathbf{Y}_{n}-\mathbf{X}_{n, \ell}$. Next, $\partial \mathcal{L}_{\mathrm{UTV}} / \partial \widehat{\mathbf{X}}_{n, \ell}$ is found by taking the derivative of (\ref{equ:UTVobj}):

\begin{equation}
\frac{\partial \mathcal{L}_{\mathrm{UTV}}}{\partial \widehat{\mathbf{X}}_{n, \ell}}=\lambda_{1}\operatorname{div}\left(\frac{\nabla_x \widehat{\mathbf{X}}_{n, \ell}}{\|\nabla_x \widehat{\mathbf{X}}_{n, \ell}\|}\right)+\lambda_{2}\operatorname{div}\left(\frac{\nabla_y \mathbf{Z}}{\|\nabla_y \mathbf{Z}\|}\right),
\end{equation}
where $\operatorname{div}(\cdot)$ is referred to as the divergence of a given matrix at a given point.

The second essential step is to focus on the calculation of $\partial \mathcal{L} / \partial \mathbf{B}_{\ell}$, which is computed in a similar way to $\delta \mathcal{W}_{\ell}$ and is defined in detail as follows:
\begin{equation}
\label{equ:LB}
\delta \mathbf{B}_{\ell}=\frac{\partial \mathcal{L}}{\partial \mathbf{B}_{\ell}}=\frac{\partial \mathcal{L}_{\mathrm{MSE}}}{\partial \widehat{\mathbf{X}}_{n, \ell}} \frac{\partial \widehat{\mathbf{X}}_{n, \ell}}{\partial \mathbf{B}_{\ell}}+\frac{\partial \mathcal{L}_{\mathrm{UTV}}}{\partial \widehat{\mathbf{X}}_{n, \ell}} \frac{\partial \widehat{\mathbf{X}}_{n, \ell}}{\partial \mathbf{B}_{\ell}}.
\end{equation}

Now, we must smoothly compute the remaining derivatives ${\partial \widehat{\mathbf{X}}_{n, \ell}}/ {\partial \mathbf{W}_{\ell-1}}$ and ${\partial \widehat{\mathbf{X}}_{n, \ell}}/ {\partial \mathbf{B}_{\ell}}$; this is specifically done by differentiating (\ref{equ:ae2}) with respect to $\mathbf{W}_{\ell-1}$ to compute ${\partial \widehat{\mathbf{X}}_{n, \ell}}/ {\partial \mathbf{W}_{\ell-1}}$ as follows.

\begin{equation}
\frac{\partial \widehat{\mathbf{X}}_{n, \ell}}{\partial \mathbf{W}_{\ell-1}}=\sigma^{\prime}\left(\mathbf{W}_{\ell-1} \mathbf{X}_{n,\ell-1}+\mathbf{b}_{\ell-1}\right)\mathbf{X}_{n,\ell-1},
\end{equation}
\textcolor{black}{where $\sigma^{\prime}$ denotes the derivative
of the activation function $\sigma$. Next, one can differentiate (\ref{equ:ae2}) with respect to the bias $\mathbf{B}_{\ell}$ and compute ${\partial \widehat{\mathbf{X}}_{n, \ell}}/ {\partial \mathbf{B}_{\ell-1}}$ as follows:}
\begin{equation}
\frac{\partial \widehat{\mathbf{X}}_{n, \ell}}{\partial \mathbf{B}_{\ell-1}}=\sigma^{\prime}\left(\mathbf{W}_{\ell-1} \mathbf{X}_{n,\ell-1}+\mathbf{b}_{\ell-1}\right).
\end{equation}

In conclusion, the above derivation is helpful for comprehending how equations (\ref{equ:LW}) and (\ref{equ:LB}) are partitioned into new UTV and MSE terms to achieve footprint removal during the alternating updates of $\partial \mathbf{X}_{n,\ell} / \partial \mathbf{W}_{\ell-1}$ and $\partial \mathbf{X}_{n,\ell} / \partial \mathbf{B}_{\ell}$. Certainly, an in-depth mathematical analysis of the above problem exceeds this article's scope. In addition, it should be noted that $\lambda_1$ and $\lambda_2$ are the most relevant parameters for the FR-Net model, and their specific optimization choices are discussed in Section~\ref{subsec:para}.

\subsection{Implementation Details}
\label{sec:implement}
This section highlights and treats two issues related to the details of the FR-Net implementation: the network architecture and the construction of unsupervised training samples. Specifically, for the $\boldsymbol{h}_{\Theta}(\cdot)$ function in equation (\ref{equ:FRobj}), one can adopt a convolutional U-Net AE as the FR-Net, which takes $\mathbf{Y}_{n,\ell}$ as its input to remove footprint noise. In this case, the nineteen-layer architecture details of U-Net are shown in Fig.~\ref{Fig:Schem}. A contracting path and an expanding path are included in this architecture; they are joined by symmetric skip connections, stride convolutions, and max pooling. Specifically, the contracting path consists of two $3\times3$ convolutions followed by a rectified linear unit (ReLU) and a $2\times2$ max pooling operation with a stride of 2 for downsampling. The expanding path performs the reverse of the contracting path process.

The second issue concerns the development of an unsupervised training sample generation process for
input 3-D seismic data. Considering that footprints appear to cross the time slices, it is natural to flatten the 3D seismic data into many time slice images. Then, we crop each time slice image $\mathbf{Y}$ into overlapping or nonoverlapping patches $\{\mathbf{Y}_n\}_{n=1}^N$ to guarantee the adequacy of the training samples. Finally, once the trained FR-Net completes the footprint removal procedure, the inverse process is executed to reorganize the images, and the final result is obtained.

\begin{figure*}[t!]

	\vspace{-0.4cm}
	\centering
	\includegraphics[width=17cm]{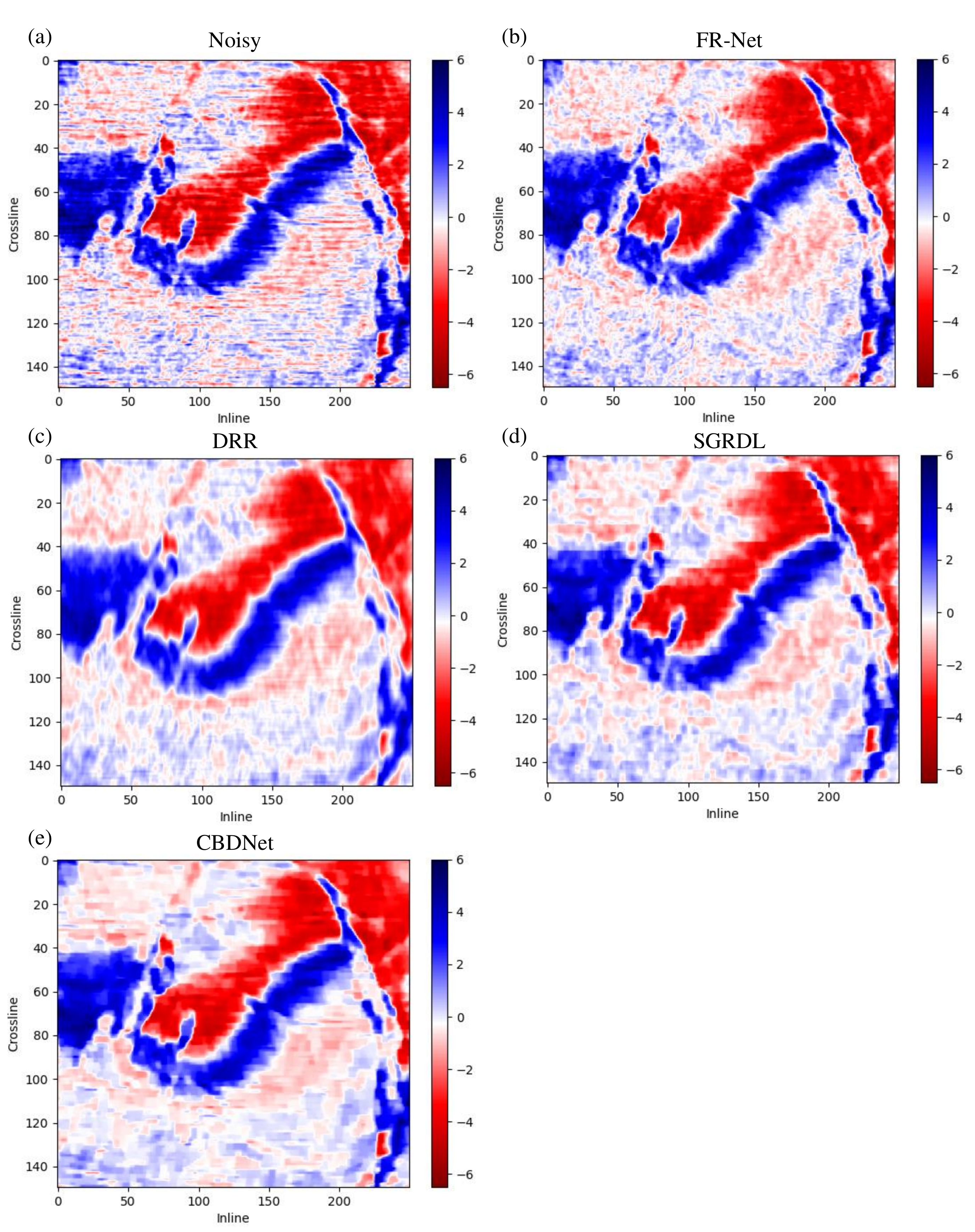}
	\caption{Results obtained on the Penobscot-3D dataset in the constant time slice comparison. (a) The constant time slice and footprint removal results are obtained using (b) the proposed FR-Net, (c) the DRR method, (d) the SGRDL method, and (e) the CBDNet method.}
\label{Fig:PenoRec} 
\end{figure*}
\begin{figure*}[t!]

	\vspace{-0.4cm}
	\centering
	\includegraphics[width=17cm]{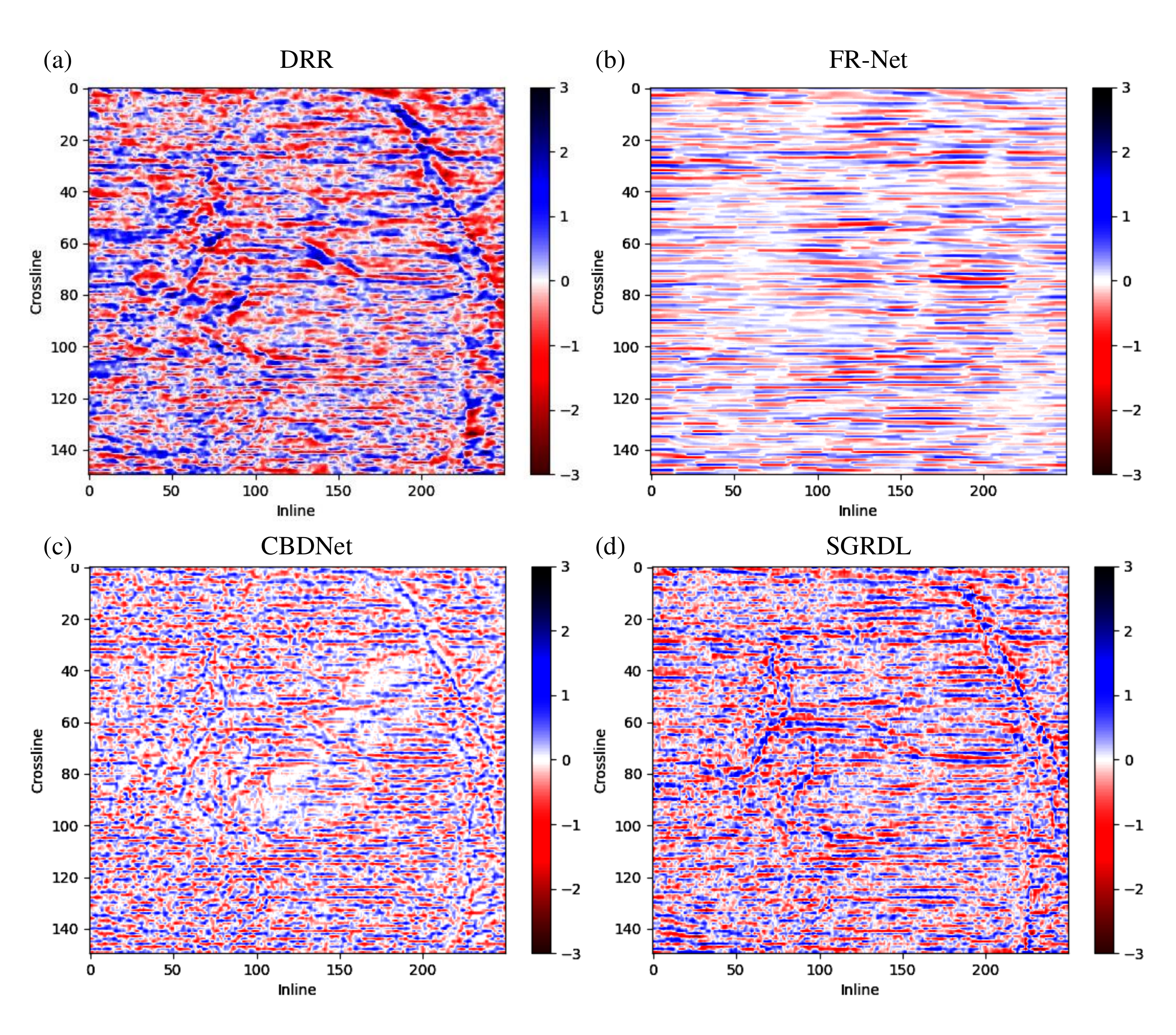}
	\caption{Results obtained on the Penobscot-3D dataset in the constant time slice comparison. The removed footprints are obtained using (a) the DRR method, (b) the proposed FR-Net method, (c) the CBDNet method, and (d) the SGRDL method.}
\label{Fig:PenobscotRemove} 
\end{figure*}

\begin{figure*}[t!]

	\vspace{-0.4cm}
	\centering
	\includegraphics[width=18cm]{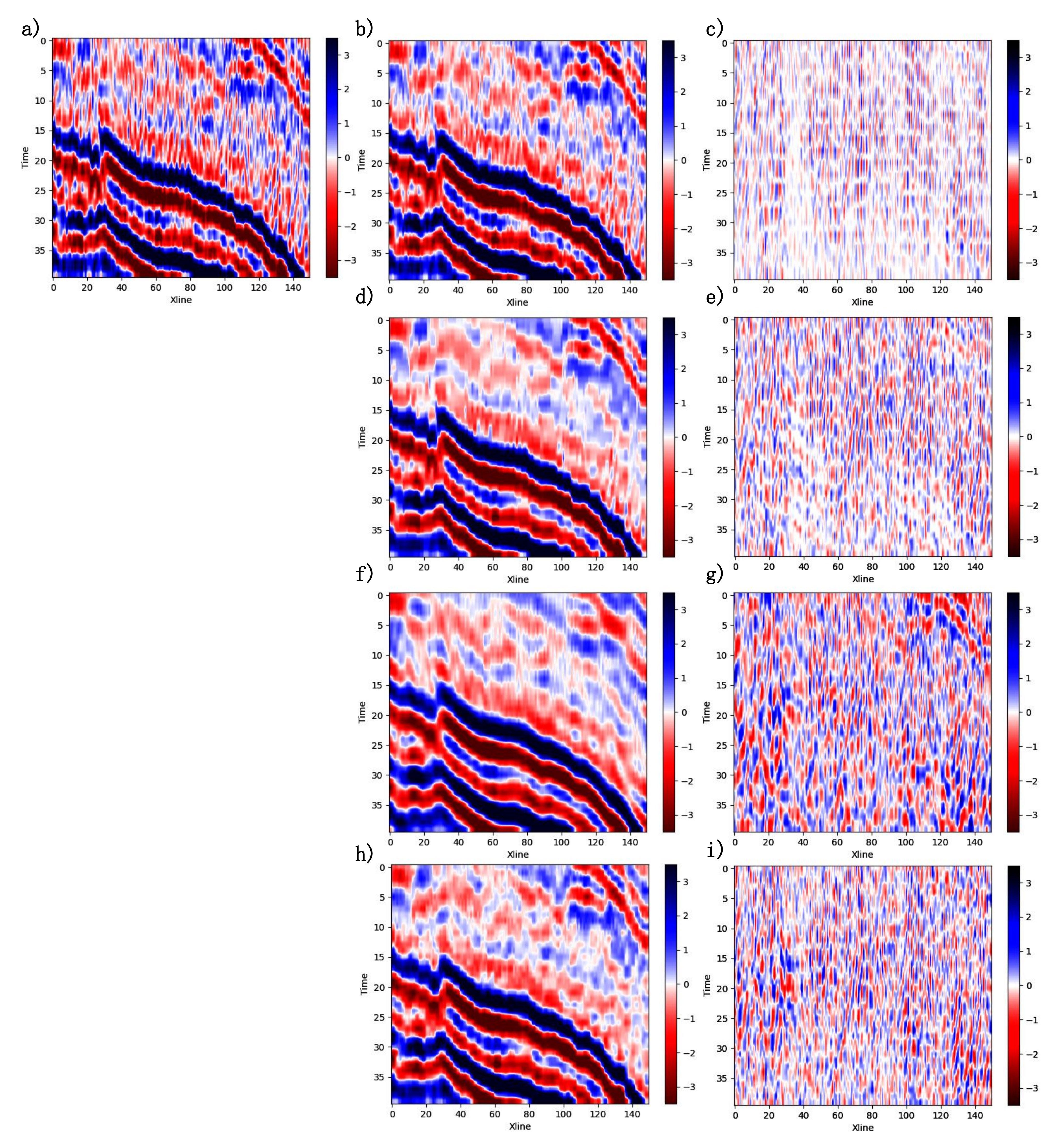}
\caption{Results obtained on the Penobscot-3D dataset in the inline section comparison. (a) Original 200th inline section. The footprint removal results are obtained using (b) the proposed FR-Net, (d) the DRR method, (f) the CBDNet method, and (h) the SGRDL method. The corresponding residuals are obtained using (c) the proposed FR-Net, (e) the DRR method, (g) the CBDNet method, and (i) the SGRDL method. }


\label{Fig:PenoRecin} 
\end{figure*}

\begin{figure*}[t!]

	\vspace{-0.4cm}
	\centering
	\includegraphics[width=17cm]{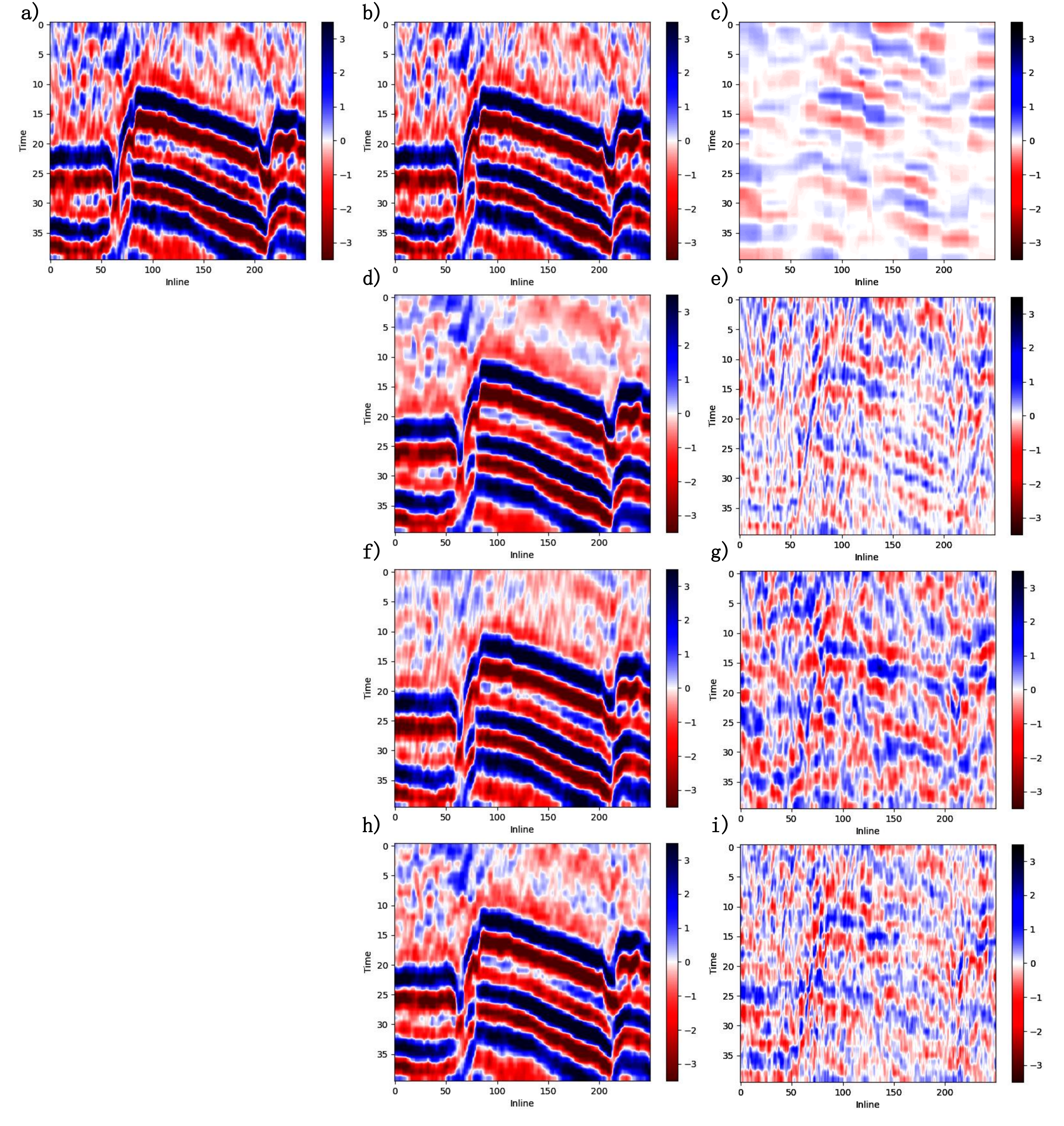}
\caption{Results obtained on the Penobscot-3D dataset in the x-line section comparison. (a) Original 40th x-line section. The footprint removal results are obtained using (b) the proposed FR-Net, (d) the DRR method, (f) the CBDNet method, and (h) the SGRDL method. The corresponding residuals are obtained using (c) the proposed FR-Net, (e) the DRR method, (g) the CBDNet method, and (i) the SGRDL method. }
\label{Fig:PenoRecinx} 
\end{figure*}

\begin{figure*}[t!]

	\vspace{-0.4cm}
	\centering
	\includegraphics[width=16.5cm]{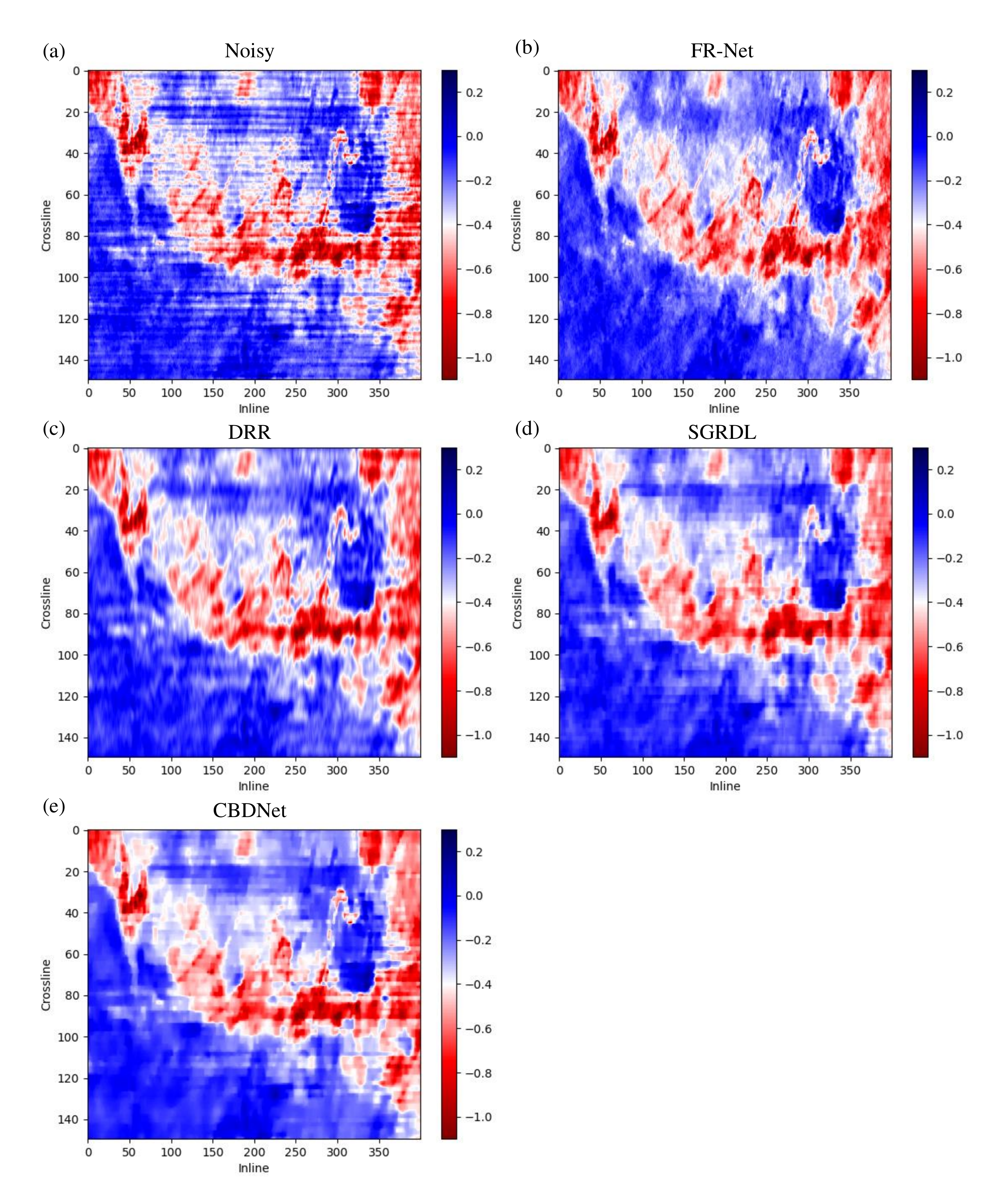}
	\caption{Results obtained on the Kerry-3D dataset in the constant time slice comparison. (a) The constant time slice and footprint removal results are obtained using (b) the proposed FR-Net, (c) the DRR method, (d) the SGRDL method, and (e) the CBDNet method.}
\label{Fig:KerryRec} 
\end{figure*}

 \begin{figure*}[t!]

	\vspace{-0.4cm}
	\centering
	\includegraphics[width=17cm]{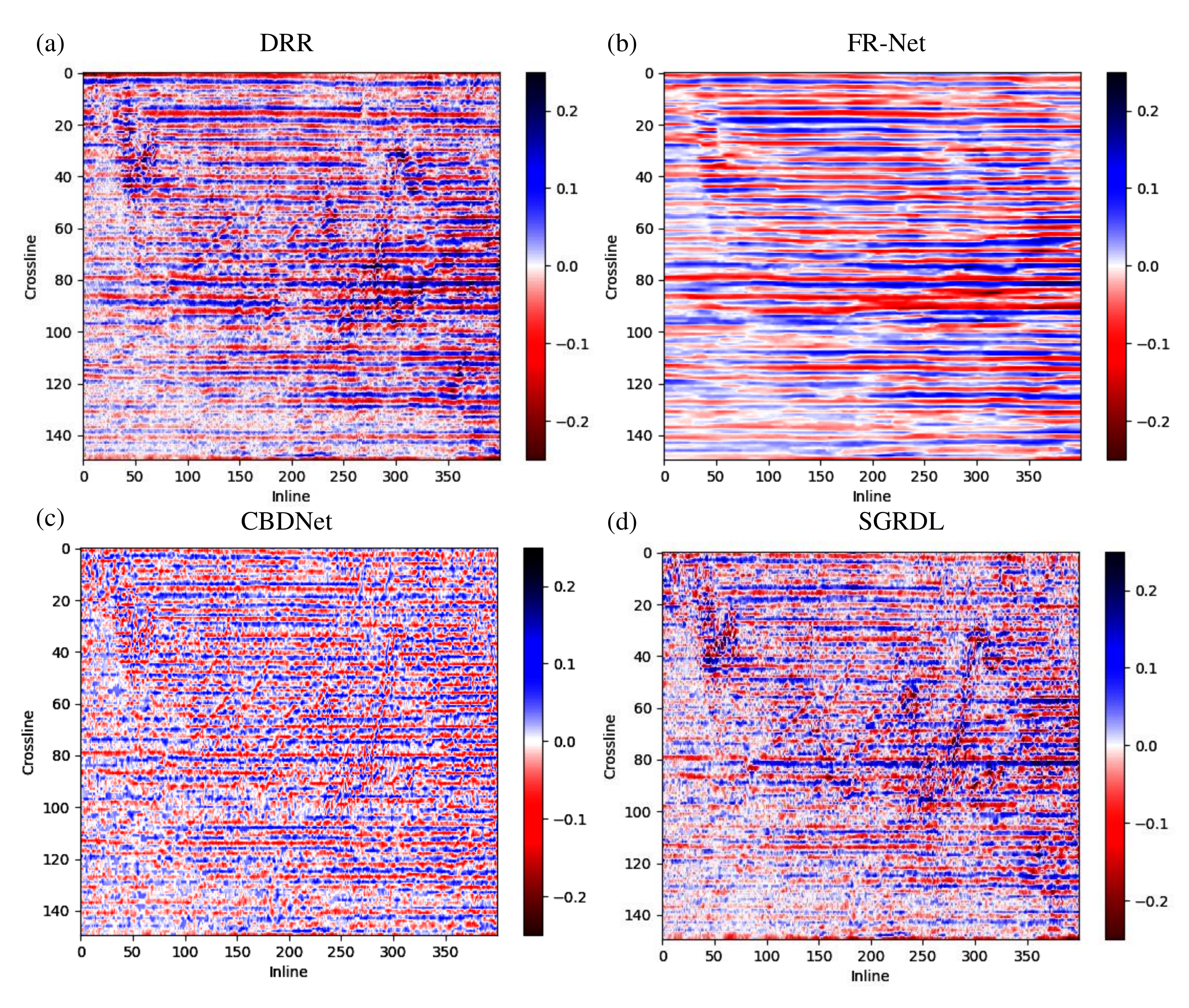}
	\caption{Results obtained on the Kerry-3D dataset in the constant time slice comparison. The removed footprints are obtained using (a) the DRR method, (b) the proposed FR-Net method, (c) the CBDNet method, and (d) the SGRDL method.}
\label{Fig:KerryErr} 
\end{figure*}

\begin{figure*}[t!]

	\vspace{-0.4cm}
	\centering
	\includegraphics[width=17cm]{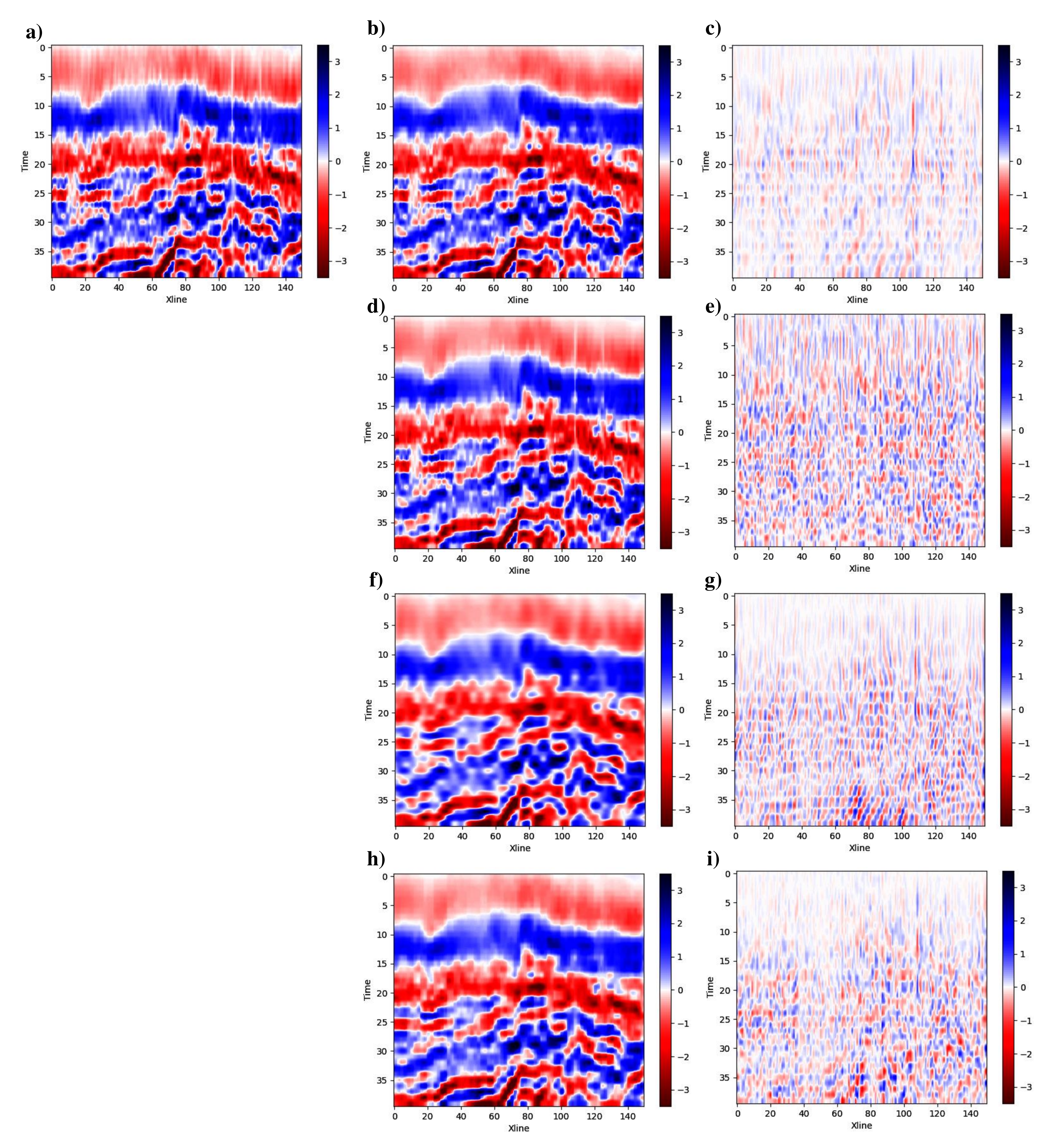}
\caption{Results obtained on the Kerry-3D dataset in the inline section comparison. (a) Original 163rd inline section. The footprint removal results are obtained using (b) the proposed FR-Net, (d) the DRR method, (f) the CBDNet method, and (h) the SGRDL method. The corresponding residuals are obtained using (c) the proposed FR-Net, (e) the DRR method, (g) the CBDNet method, and (i) the SGRDL method. }
\label{Fig:KerryRecIn} 
\end{figure*}

\begin{figure*}[t!]

	\vspace{-0.4cm}
	\centering
	\includegraphics[width=18cm]{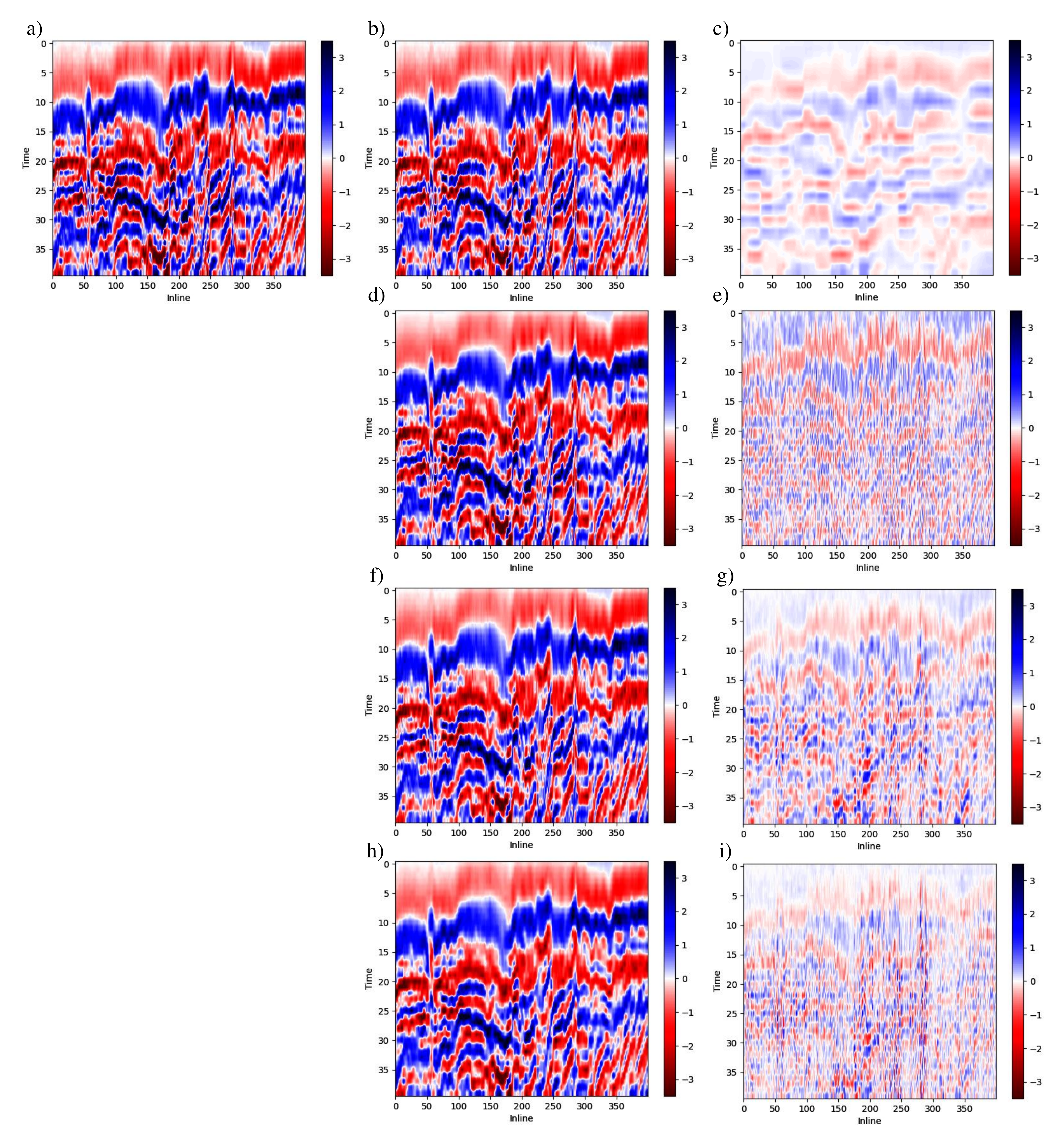}
\caption{Results obtained on the Kerry-3D dataset in the x-line section comparison. (a) Original 70th x-line section. The footprint removal results are obtained using (b) the proposed FR-Net, (d) the DRR method, (f) the CBDNet method, and (h) the SGRDL method. The corresponding residuals are obtained using (c) the proposed FR-Net, (e) the DRR method, (g) the CBDNet method, and (i) the SGRDL method. }
\label{Fig:KerryRecX} 
\end{figure*}

\section{Experimental Results}
\label{Sec:examp}

Once our FR-Net model build is implemented, it is time to ascertain the efficacy of such a model through a series of comparison experiments conducted on both synthetic and real field data. First, the experimental configuration consists of the utilized baselines, datasets, performance metrics, and parameter settings, which are presented in Sections~\ref{subsec:base}--\ref{subsec:para}, respectively. All experiments are divided into two parts: one part uses synthetic data in Section~\ref{subsec:syn}, and the other part uses real field data in Section~\ref{subsec:Penobscot}. For a related discussion concerning the experiments, see Section~\ref{subsec:disc}.

\subsection{Baseline Algorithms}
\label{subsec:base}
Since our proposed FR-Net has both classic DCAE and UTV components, we compare our proposed method against the classic TV-regularized DCAE model~\cite{guo2019toward} to verify the benefits provided by the use of UTV. According to the seismic literature review in Section~\ref{sec:Intro}, we also compare our approach with the recent footprint removal model~\cite{chen2021statistics} and SVD-based filtering~\cite{chen2016simultaneous}, which learns how to efficiently eliminate noise from contaminated data, as done in~\cite{chen2021statistics}. These SOTA methods are detailed as follows.

\begin{enumerate}
\item \textit{Convolutional blind denoising network
(CBDNet)~\cite{guo2019toward}\footnote{\url{https://github.com/GuoShi28/CBDNet}}:} For a fair comparison, both our FR-Net and the CBDNet model share the same network architecture but use different TV regularity terms; i.e., the CBDNet includes conventional TV, while the FR-Net includes UTV. This comparison helps to highlight the benefits of our UTV penalty term for footprint removal.

\item \textit{Statistics-guided residual dictionary learning (SGRDL)~\cite{chen2021statistics}\footnote{\url{https://github.com/chenyk1990/reproducible_research/tree/master/foot}\label{foot:sgrdl}}:} SGRDL generates an algorithmic framework for learning the dictionary atoms of signal waveforms and isolating the footprint characteristics from the learned atoms. In this instance, SGRDL utilizes an approach based on statistical analysis for classifying dictionary atoms as footprint-affected or footprint-free.

\item \textit{Damped rank reduction (DRR)~\cite{chen2016simultaneous}\textsuperscript{\ref{foot:sgrdl}}:} By introducing a damping operator into the conventional truncated SVD model, DRR tends to partition the vector space of the Hankel matrix of the noisy signals into a useful signal subspace and a noise subspace.

\end{enumerate}

In the experiment, DRR and SGRDL are implemented in MATLAB, while the CBDNet and FR-Net are realized in TensorFlow\footnote{\url{https://www.tensorflow.org/}} on the Google Colab platform\footnote{\url{https://colab.research.google.com/}}; keep in mind that Colab assigns GPUs to users at random on the platform. In addition, for all of the experiments, the model parameters of DRR and SGRDL are carefully adjusted to reach performance comparable to that reported in~\cite{chen2016simultaneous} and~\cite{chen2021statistics}.

\subsection{Testing Datasets}
\label{subsec:data}
This section seeks to exemplify the superior performance of the proposed FR-Net in three comparisons, one of which uses synthetic field data while the others use actual field data. Each piece of data used in this article is freely accessible to the public and is described in detail below.
\begin{enumerate}
\item \textit{Synthetic data\textsuperscript{\ref{foot:sgrdl}}:} This 3-D synthetic dataset, used by~\cite{chen2021statistics}, simply consists of three linear seismic events, each of which has a different wavelet. The clear data deliberately add a severe footprint to produce the final 3D observations.
\item \textit{Penobscot-3D\footnote{\url{https://wiki.seg.org/wiki/Penobscot_3D}}:} Kington's Penobscot subset gives researchers access to a publicly available dataset with footprint patterns that are visible on shallow time slices. Each footprint pattern consists of a succession of stripes with different lengths that follow the direction of a crossline.
\item \textit{Kerry-3D\footnote{\url{https://wiki.seg.org/wiki/Kerry-3D}}:} The upper portion of the Kerry-3D dataset is extensively diluted by footprint noise resulting from horizontal bands with varied widths and lengths; this dataset is far more polluted than Penobscot.
\end{enumerate}

These datasets are sufficient as benchmarks for comparing the footprint removal algorithms and for comprehensively illuminating the features of the FR-Net because they vary in difficulty from easy to tough.

\subsection{Performance Metric}
\label{subsec:perf}
In the synthetic data experiments, the ground-truth data are available and can be compared with the footprint removal results. Apart from visual comparisons, it is quite natural to use quantitative evaluation metrics to compare the denoising performance of the FR-Net and SOTA methods. Inspired by~\cite{chen2021statistics}, we also use the signal-to-noise ratio (SNR) as a quantitative evaluation metric.

\begin{equation}
\mathrm{SNR[dB]}=10 \log _{10} \frac{\|\mathbf{X}\|_{2}^{2}}{\|\mathbf{X}-\widehat{\mathbf{X}}\|_{2}^{2}},
\label{equ:snr}
\end{equation}
where $\mathbf{X}$ and $\widehat{\mathbf{X}}$ denote the clean image and the estimated image, respectively.

\subsection{Parameter Settings}
\label{subsec:para}
After the FR-Net architecture in Section~\ref{sec:implement} is arbitrarily chosen, the next task is to preset two categories of parameters: the parameters of the cost function and the training algorithm parameters. The cost function of the FR-Net involves two explicit tradeoff parameters that must be chosen: $\lambda_1$ and $\lambda_2$. Here, we set $\lambda_1=6 \times 10^{4}$ and $\lambda_2=4 \times 10^{5}$ for the synthetic data, $\lambda_1=4.5 \times 10^{4}$ and $\lambda=3\times 10^{5}$ for the Penobscot data, and $\lambda_1=5 \times 10^{4}$ and $\lambda_2=3 \times 10^{5}$ for the Kerry-3D data. Once the parameters of the cost function are known, hyperparameter optimization (i.e., with the Adam optimizer) is performed to select the optimal training algorithm parameters, such as the batch size and learning rate. Specifically, the final learning rate is set to $1.0 \times 10^{-5}$ for both the synthetic and Kerry-3D data and to $1.0 \times 10^{-4}$ for the Penobscot data, and the batch size is set as 5 for all datasets.

\subsection{Synthetic Data Experiments}
\label{subsec:syn}
To validate the FR-Net model, we test it on a 3-D synthetic dataset created by the open source code in~\cite{chen2021statistics} with a deliberately added vertical footprint that decays with the inline, as plotted in Fig.~\ref{Fig:syn1}.  Notably, the FR-Net does not see clean seismic images at all during the footprint removal process because it exists only to calculate SNR values as it did earlier. Actually, our preprocessing stage must split the noisy seismic images into $50 \times 50$ patches, where all patches can be fed to the FR-Net.

Figs.~\ref{Fig:syn2} and ~\ref{Fig:Syn3} show the results of two constant slice comparisons  conducted on the synthetic datasets in terms of the obtained recovery effects and residuals. From Figs.~\ref{Fig:syn2} and~\ref{Fig:Syn3}, it is also confirmed that the FR-Net leaves almost no residual footprint, while the SOTA methods cannot completely eliminate all footprint noise. Furthermore, the footprint residuals of the SOTA and FR-Net methods can be proven more intuitively by a single-trace comparison, as shown in Fig.~\ref{Fig:trace}. \textcolor{black}{It is worth noting that the FR-Net results in Fig.~\ref{Fig:trace} are nearly superb and keep the useful signal completely separate from footprint-contaminated seismic data, thanks to the fact that all simulation footprints attempt to strictly satisfy the UTV directional assumptions.} Apart from visual comparisons, a quantitative SNR comparison between them is examined to characterize the residual footprint levels, as shown in Table~\ref{table:SNR}. From Table~\ref{table:SNR}, our FR-Net model produces the smallest residual footprint (36.00 $\mathrm{dB}$) relative to those of the DRR, CBDNet, and SGRDL techniques, which achieve values of 10.29, 20.69, and 20.10 $\mathrm{dB}$, respectively. \textcolor{black}{As observed, it is apparent that the proposed method seamlessly decouples the valid reflected signal from noisy seismic images without any loss of information.}

\subsection{Real Field Data Experiments}
\label{subsec:Penobscot}
To verify the superiority and effectiveness of the proposed method in practice, the FR-Net is applied to two distinct real field datasets: Penobscot-3D and Kerry-3D. As stated in Section~\ref{subsec:data}, these real field data cover an extensive range of footprint noise levels, which assists in ascertaining the adaptability of the proposed FR-Net to a variety of noise intensities. Thus, depending on the selected footprint noise level, the entire real field data test currently consists of the following two subexperiments.


In the first example, we demonstrate the efficacy and accuracy of the FR-Net on the Penobscot-3D dataset. The Kerry-3D dataset possesses a large acquisition footprint, which can involve significant engineering challenges for the Penobscot data, as shown in Fig.~\ref{Fig:PenoRec}(a). In detail, Figs.~\ref{Fig:PenoRec}(b)-(e) and Figs.~\ref{Fig:PenobscotRemove}(a)-(d) show the noise cancellation outcomes and corresponding footprint removal effects achieved by employing the proposed FR-Net method and the SOTA methods. From Fig.~\ref{Fig:PenoRec} and Fig.~\ref{Fig:PenobscotRemove}, it is clear that the FR-Net can effectively remove the footprint without degrading the seismic image resolution, while the SOTA algorithms tend to either produce oversmoothed image edges or leave residual footprints in the resulting image. To further compare the footprint removal effects, two lines in Fig.~\ref{Fig:PenoRec}(a) are selected for section inspections, as shown in Figs.~\ref{Fig:PenoRecin} and ~\ref{Fig:PenoRecinx}. Undoubtedly, all results in Figs.~\ref{Fig:PenoRecin} and ~\ref{Fig:PenoRecinx} replicate the initial findings and therefore corroborate the conclusion that the FR-Net method achieves significant footprint attenuation where no signal leakage occurs, and the edges are effectively preserved.

 

In the second example, we use Kerry-3-D field data with serious footprint noise to further test our proposed FR-Net. As before, it is natural to first compare the four methods on a time slice, as shown in Figs.~\ref{Fig:KerryRec} and~\ref{Fig:KerryErr}. Regarding visual quality, the proposed FR-Net produces the closest result to the original seismic image except for the removal of footprint noise, making it superior to the other SOTA methods. Similarly, Figs.~\ref{Fig:KerryRecIn} and~\ref{Fig:KerryRecX} show the two lines selected from Fig.~\ref{Fig:KerryRec}(a) to further investigate the footprint noise removal performance of the compared methods. From Figs.~\ref{Fig:KerryRecIn} and~\ref{Fig:KerryRecX}, it can be observed that the FR-Net deletes fewer meaningful signals in the footprint residual, while the SOTA approaches always cause too much damage to the useful signals. This leads us to conclude that the proposed FR-Net approach can greatly improve upon the footprint removal results of the SOTA methods for Kerry-3-D data with heavy footprint noise.




\subsection{Discussion}
\label{subsec:disc}
As in~\cite{chen2021statistics},~\cite{liu2021dictionary},~\cite{gomez2020footprint}, it is assumed that useful signals, footprints, and random noise are additive. In this case, we train an FR-Net to separate useful signals from footprints and concentrate them into the corresponding independent components. From Figs.~\ref{Fig:syn2},~\ref{Fig:Syn3},~\ref{Fig:PenoRec}, and~\ref{Fig:KerryRec}, it is quite evident that the footprints are removed very cleanly, but inevitably, much random noise remains in the useful signals, which is the principal drawback of our FR-Net. The root cause is the problem that the pure U-Net fails to effectively eliminate random noise in an unsupervised learning manner~\cite{guo2019toward}. Notably, for the classic noise reduction task, various methods have been introduced to compensate for this deficiency, such as structure-oriented filtering~\cite{fehmers2003fast}.

In terms of computational complexity, our FR-Net is almost the same as the standard U-Net because we only add a regular UTV term to its cost function and do not change the network architecture or incur any inference overhead. As typical DL networks, both U-Net and the FR-Net also suffer from high computational complexity due to the recursive nature of the training procedure, which results in a common issue for any DL
model.

As with the standard U-Net, the FR-Net is also parameter-sensitive, and it would require exhaustive experiments to obtain the optimal parameters. In addition to these network parameters, such as the training rate and batch size used by the standard U-Net, it is essential for the FR-Net to adjust the tradeoff values $\lambda_1$ and $\lambda_2$ of equation (\ref{equ:tdlobj}), which significantly affect the consequences of footprint removal. Specifically,
we must set the tradeoff parameters $\lambda_1$ and $\lambda_2$ to appropriate values that are neither too low nor too high. If $\lambda_1$ and $\lambda_2$ are too large, the FR-Net will remove too much footprint noise and distort the underlying signal. If the tradeoff parameters $\lambda_1$ and $\lambda_2$ are too small, the FR-Net will be inefficient.


\begin{table}[t]
\renewcommand{\arraystretch}{1.3}
\centering
\caption{Quantitative Comparison Among the Footprint Removal Results of DRR, the CBDNet, SGRDL, and the Proposed FR-Net Method on Synthetic Datasets}  
\label{table:SNR} 
\centering
\begin{center}  
\begin{tabular}{|l|c|}  
\hline   
{Method}  & {Synthetic Dataset}    \\    
\hline\hline   
{DRR}  & $10.29~\mathrm{dB}$     \\ 
  \hline   
{{CBDNet}} & ${20.69}~\mathrm{dB}$     \\ 
\hline  
{SGRDL}  & $\text{20.10}~\mathrm{dB}$      \\ 
  \hline   
{Proposed FR-Net} & $\textbf{36.00}~\mathrm{dB}$     \\ 
\hline  
\end{tabular}  
\end{center} 
\end{table}

\section{Conclusion}
\label{Sec:conclusion}
This article proposes a \textcolor{black}{physical prior augmented} footprint removal network, namely, the FR-Net, to suppress acquired footprints in an unsupervised way. Specifically, to capture the intrinsically directional properties of footprints, we design a UTV model, which is quite distinct from all current removal methods. Then, strongly regularizing the DL method using the UTV model produces an ultimate FR-Net, which transforms DL from an entirely data-driven approach to a \textcolor{black}{physical prior augmented} technique. By jointly optimizing the FR-Net model in an unsupervised manner, the complete separation of footprint noise and useful signals is projected by the BP algorithm.
Experimental results obtained on synthetic and real field datasets demonstrate the substantial gains achieved by our method over the SOTA methods in terms of their SNR valuations and visual quality.

Future work in this area will continue to extend the proposed \textcolor{black}{physical prior augmented} DL framework to other types of noise attenuation, e.g., swell noise, ground-roll noise, and desert noise. \textcolor{black}{We will also explore the combination of UTV and other models, for example, low-rank approximation (LRA), sparse representation (SR), and dictionary learning.  In this case, we not only successfully decoupled the footprint with the help of UTV but also nurtured some unforeseen properties, such as the ability to compensate for FR-Net defects that fail to suppress the random noise.}

\section*{\textcolor{black}{Acknowledgments}}
\textcolor{black}{The authors would like to thank the associate editor, and the two anonymous reviewers, whose constructive suggestions helped to greatly improve and clarify this article.}

\ifCLASSOPTIONcaptionsoff
  \newpage
\fi

\end{document}